\documentclass[lettersize,journal]{IEEEtran}
\usepackage[misc]{ifsym}
\usepackage{amsmath,amsfonts}
\usepackage{array}
\usepackage{hyperref}
\usepackage[caption=false,font=normalsize,labelfont=sf,textfont=sf]{subfig}
\usepackage{textcomp}
\usepackage{stfloats}
\usepackage{url}
\usepackage{verbatim}
\usepackage{graphicx}
\usepackage{times}
\usepackage{epsfig}

\usepackage{amsmath}
\usepackage{amssymb}
\usepackage{pmat}
\usepackage[noend]{algpseudocode}
\usepackage{algorithmicx,algorithm}
\usepackage{multirow}
\usepackage{color}
\usepackage{booktabs}

\usepackage{makecell}
\usepackage{xcolor}
\usepackage{extarrows}

\newcolumntype{M}[1]{>{\centering\arraybackslash}m{#1}}
\newcommand{\eg}{\emph{e.g. }{}}

\ifCLASSOPTIONcompsoc
  \usepackage[nocompress]{cite}
\else
  \usepackage{cite}
\fi

\begin{document}
%
\title{GUPNet++: Geometry Uncertainty Propagation Network for Monocular 3D Object Detection. }

\author{Yan~Lu, Xinzhu~Ma, Lei~Yang, Tianzhu~Zhang,~\IEEEmembership{Member,~IEEE}, Yating~Liu, Qi~Chu, \\
Tong~He$^\text{\Letter}$, Yonghui~Li,~\IEEEmembership{Fellow,~IEEE}, Wanli~Ouyang,~\IEEEmembership{Senior~Member,~IEEE}
\thanks{Yan~Lu and Yonghui~Li are with the School of Electrical and Information Engineering, University of Sydney, NSW, Australia (e-mail: yalu6572@uni.sydney.edu.au and yonghui.li@sydney.edu.au).
Yan~Lu, Xinzhu~Ma, Tong~He and Wanli~Ouyang are with Shanghai AI Laboratory, Shanghai, China (e-mail: \{luyan, maxinzhu, hetong, ouyangwanli\}@pjlab.org.cn). 
Lei~Yang is with SenseTime Group Limited, Shenzhen, China (e-mail: yanglei@sensetime.com) and this work was done during Yan's internship at Shanghai Artificial Intelligence Laboratory. Tianzhu~Zhang, Yating~Liu and Qi~Chu are with University of Science and Technology of China, Hefei, China (e-mail: \{tzzhang, qchu\}@ustc.edu.cn.com and liuyat@mail.ustc.edu.cn).}
\thanks{Manuscript received April 19, 2021; revised August 16, 2021.}}
\markboth{Journal of \LaTeX\ Class Files,~Vol.~14, No.~8, August~2021}%
{Shell \MakeLowercase{\textit{et al.}}: A Sample Article Using IEEEtran.cls for IEEE Journals}


\maketitle
\renewcommand{\thefootnote}{\Letter}
\footnotetext{Corresponding author.}
\renewcommand{\thefootnote}{1}

\begin{abstract}
Geometry plays a significant role in monocular 3D object detection. It can be used to estimate object depth by using the perspective projection between object's physical size and 2D projection in the image plane, which can introduce mathematical priors into deep models. However, this projection process also introduces error amplification, where the error of the estimated height is amplified and reflected into the projected depth. It leads to unreliable depth inferences and also impairs training stability. 
To tackle this problem, we propose a novel Geometry Uncertainty Propagation Network (GUPNet++) by modeling geometry projection in a probabilistic manner. This ensures depth predictions are well-bounded and associated with a reasonable uncertainty. 
The significance of introducing such geometric uncertainty is two-fold: (1). It models the uncertainty propagation relationship of the geometry projection during training, improving the stability and efficiency of the end-to-end model learning. (2). It can be derived to a highly reliable confidence to indicate the quality of the 3D detection result, enabling more reliable detection inference.
Experiments show that the proposed approach not only obtains (state-of-the-art) SOTA performance in image-based monocular 3D detection but also demonstrates superiority in efficacy with a simplified framework.
The code and model will be released at \href{https://github.com/SuperMHP/GUPNet\_Plus}{\textcolor[RGB]{0 206 209}{https://github.com/SuperMHP/GUPNet\_Plus}}.
\end{abstract}


\section{Introduction}\label{sec:introduction}
\IEEEPARstart{M}{\lowercase{onocular}} 3D object detection is drawing increasing attention due to its application potential and low cost. 
Compared with LiDAR/stereo-based methods~\cite{meyer2019lasernet,qi2019deep,qin2019triangulation,shi2019pointrcnn,shi2019part,yang2019std,yuan2021temporal}, this task is still challenging due to the lack of depth cues, which makes object-level depth estimation naturally ill-posed. 
To this end, numerous research efforts~\cite{qin2019monogrnet,weng2019monocular,bao2020object,barabanau2019monocular,cai2020monocular,ku2019monocular} have sought to introduce geometric inductive prior to help object-wise depth estimation, among which geometry perspective projection is prevalent. 
\begin{figure}[t]
\begin{center}
\includegraphics[width=1.0\linewidth]{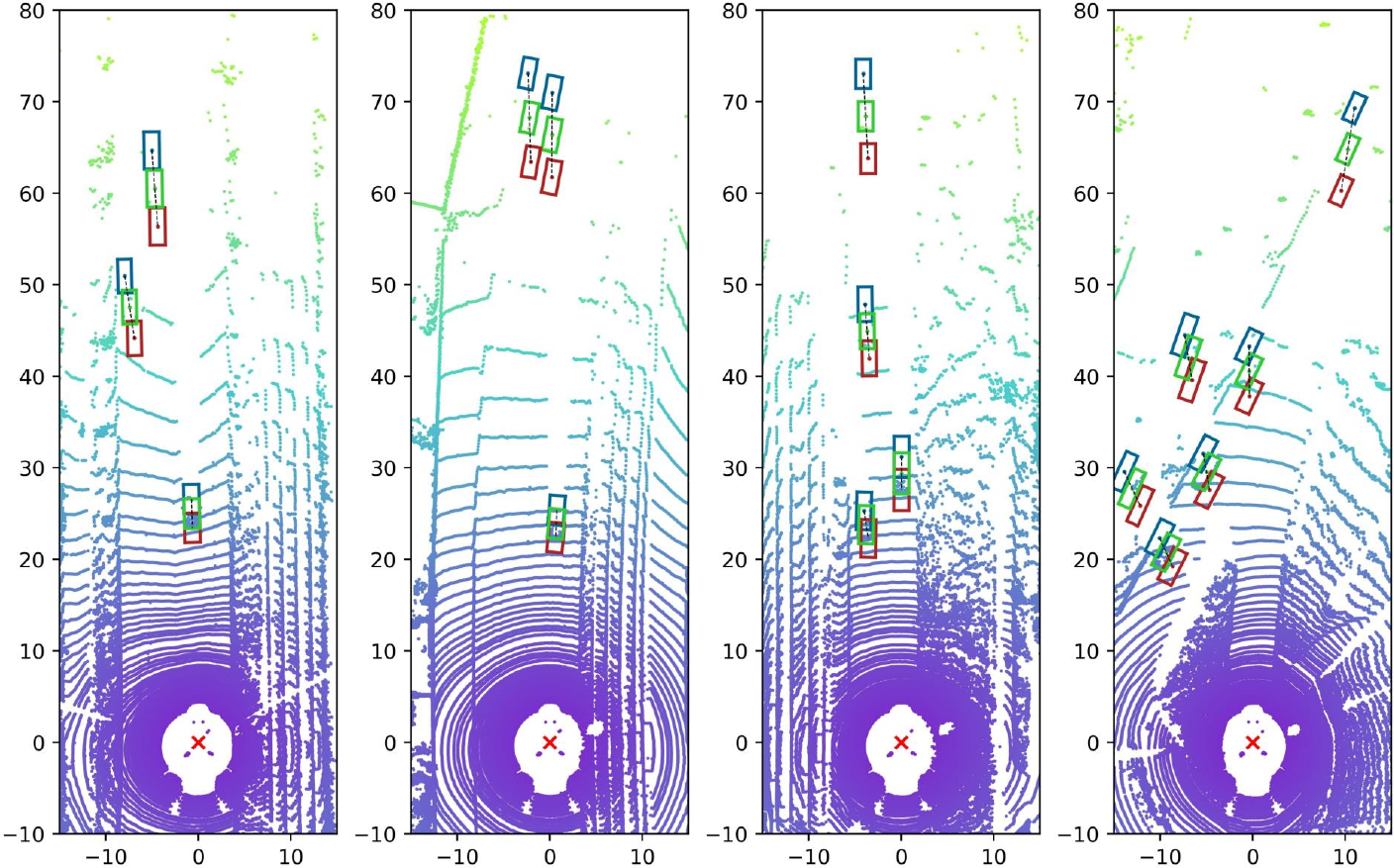}
\end{center}
   \caption{Visualized examples of depth shift caused by $\pm$0.1m 3D height jitter. We draw some bird's view examples to show the error amplification effect. The unit of the horizontal axis and the vertical axis are both meters, and the vertical axis corresponds to the depth direction. The \textcolor[RGB]{50,205,50}{green} boxes mean the original projection outputs. The \textcolor[RGB]{0,102,153}{
  blue} and \textcolor[RGB]{178,34,34}{red} boxes are shifted boxes caused by +0.1m and -0.1m 3D height bias respectively (best viewed in color). }
\label{fig:shift}
\end{figure}
\begin{figure}[t]
\begin{center}
\includegraphics[width=1.0\linewidth]{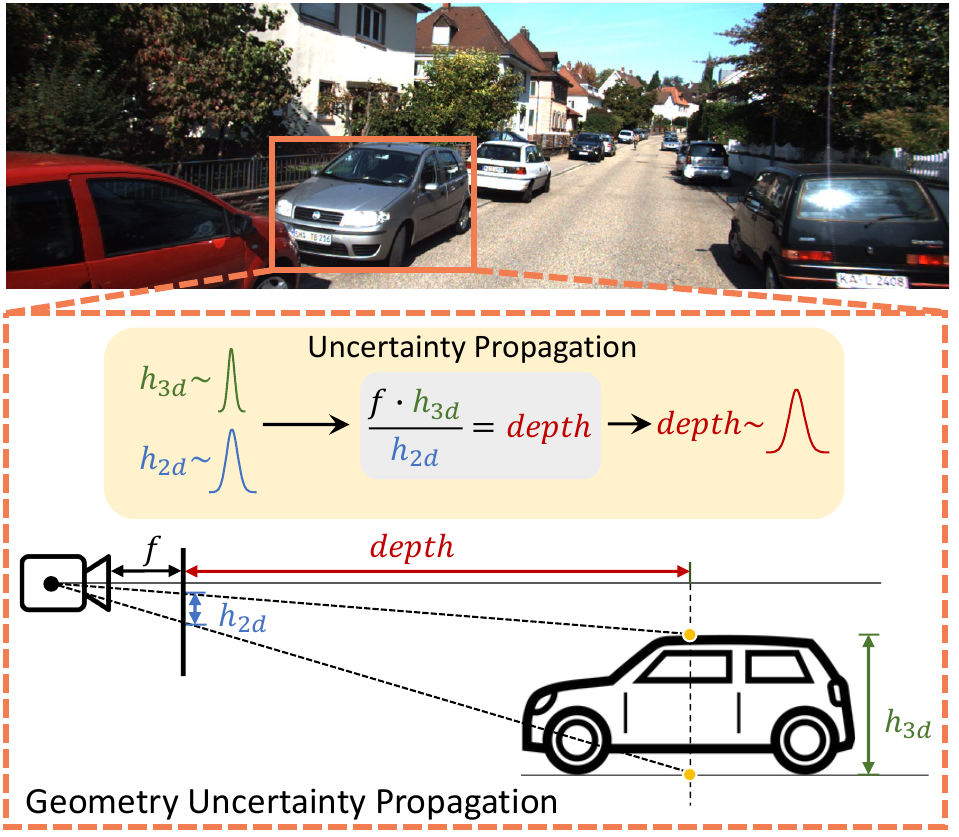}
\end{center}
   \caption{The main pipeline of our Geometry Uncertainty Propagation module. The projection process is modeled by the uncertainty theory in the probability framework. The inference depths can be represented as a distribution to provide both accurate values and scores.}
\label{fig:upm}
\end{figure}

Geometry perspective projection builds a connection among the depth, physical scale, and the size captured within images. Depth estimation can be achieved with two steps: 
1). Estimating the object's visual height on the image plane (defined as 2D height, $h_{2d}$) and physical height, also called 3D height, $h_{3d}$, respectively. 2). Inferring depth via the following formula: $depth=h_{3d}\cdot f/h_{2d}$ where $f$ is the camera focal length. 
This projection process introduces inductive prior and leads to better depth estimation results. Although promising, 
error of estimated height is also reflected in the final projected depth.  
To show the influence of this property clearly, we give examples of depth shifts caused by a fixed 3D height error in Figure~\ref{fig:shift}.
We find that even a slight drift (0.1m) of 3D height could cause a significant shift (even 4.0m) in projected depth. 
A similar phenomenon can also be found when operating perturbation on 2d height. 
This error amplification problem makes outputs of the projection-based methods hardly controllable.
During inference, it is hard for the model to accurately measure the quality of the uncontrollable depth projection and provide a reliable score for detection results, leading to the unreliable model. 
%
%
And during training,
2D/3D height estimations tend to be noisy, especially at the beginning of the training phase, making the model hard to converge and degrading the final performance, due to the exaggerated depth estimation.
To address these problems, we propose a Geometry Uncertainty Propagation Network (GUPNet++). 

Unlike standard projection-based methods, the GUPNet++  establishes the projection process in a probabilistic manner, as illustrated in Figure~\ref{fig:upm}. 
Depth is formulated as a distribution and derived from two height distributions following the geometry constraint.
The probabilistic modeling of the project process ensures that depth distribution contains statistics propagated from estimated height.
With this, a geometry-guided depth uncertainty is obtained from the depth distribution to reflect estimation error for each projected depth. 
%
During inference, this depth uncertainty
could be used to derive confidence for the 3D detection to improve the reliability of our model. Additionally, during training, we can exploit the depth uncertainty to avoid the influence of the extremely noisy projection depth values, leading our model to a more stable training process. To achieve these goals, we propose an IoU-guided Uncertainty-Confidence scheme that efficiently transfers depth uncertainty into 3D detection scores. Additionally, an Uncertainty-based Optimization method is introduced to leverage height and depth uncertainty for stabilizing the training process effectively. These techniques reduce the impact of error amplification in both training and inference phases, leading to more reliable 3D detection results.
%
%
In summary, the key contributions of this paper are as follows:
\begin{itemize}
    \item We introduce a Geometry Uncertainty Propagation Network (GUPNet++) and improve the performance of projection-based monocular 3D object detection, achieving state-of-the-art results.
    \item We develop a new geometry-guided depth uncertainty combining both mathematical priors and uncertainty modeling to overcome error amplification in the monocular 3D object detection. A novel IoU-guided Uncertainty-Confidence is proposed to utilize the uncertainty to achieve more reliable 3D detection. An Uncertainty-based Optimization method is introduced to stabilize the model training effectively.
    \item Evaluation on the challenging KITTI dataset shows the overall proposed GUPNet++ achieves state-of-the-art performance for the car on the KITTI testing set, which brings around 4.88\%, 2.28\% and 2.81\% gains on Easy, Moderate and Hard metrics, respectively than the original conference version GUPNet. Except that, evaluation on the nuScenes dataset shows that the GUPNet++ achieves a new state of the art about 34.8\% mAP. 
\end{itemize}

\noindent
{\bf Difference from the conference paper.} The preliminary version of this manuscript is the Geometry Uncertainty Network (GUPNet)~\cite{lu2021geometry}. Compared with the conference version, the improvements of the GUPNet++ are listed below:
1). We derive a new uncertainty for the projected depth in the GUPNet++. It reflects uncertainty propagated from both 2D and 3D heights to depth, which can lead to a simplified training pipeline and better inference reliability. 
2). The proposed new uncertainty model allows us to train the model directly in an uncertainty-based optimization scheme. The loss functions for the projected depth and heights are both set as uncertainty loss, which naturally stabilizes the training for projection. The new method is more flexible and free from a heuristic learning scheme, which is required by the original GUPNet.
3). We propose a new score computation algorithm, IoU-guided Uncertainty-Confidence to transfer the uncertainty into the detection score. This novel confidence metric takes into account various object properties, including dimension and orientation, providing a more reliable indicator than vanilla Uncertainty-Confidence scheme in the GUPNet. 
4). We conduct additional experiments to evaluate the effectiveness of the GUPNet++. Experiments on the nuScenes benchmark show that the GUPNet++ achieves new state-of-the-art monocular 3D detection results about 34.8\% mAP on the nuScenes test set, demonstrating the generalization of our method on large-scale datasets. Further, on the KITTI benchmark, the GUPNet++ provides significant gains over the GUPNet by a large margin, about 2.28\% on car moderate metric.

\section{Related works}

\subsection{Uncertainty in computer vision.} The uncertainty theory is widely used in deep regression~\cite{blundell2015weight}, which can model both aleatoric and epistemic uncertainty~\cite{kendall2017uncertainties}. Several computer vision methods utilized uncertainty to acquire reliable results. For example, in the 2D object detection, Softer-nms~\cite{he2018softer} proposed to train the bounding box regression in an uncertainty optimization framework, where each box has a regression uncertainty that can be used for more accurate NMS results. Similar ideas also were developed in the human pose estimation. Gundavarapu~\emph{et al.}~\cite{gundavarapu2019structured} proposed to predict multi-variate uncertainty for human joints by the structured uncertainty theory~\cite{dorta2018structured}. RLE~\cite{li2021human} went further for the joint uncertainty computation by utilizing the normalizing flow~\cite{rezende2015variational} to estimate the likelihood distribution and break the Gaussian assumption. In the image retrieval, Yu~\emph{et al.}~\cite{yu2019robust} and PCME~\cite{chun2021probabilistic} proposed to model sample embedding as a distribution rather than a fixed feature vector, treating the ambiguity property in the retrieval task well. Jin~\emph{et al.}~\cite{jin2020uncertainty} utilized uncertainty to help the single-shot retrieval system distill knowledge from the multi-shot model more reliably. Further, Kendall~\emph{et al.}~\cite{kendall2018multi} provided that uncertainty also can tackle multi-task learning. They introduced uncertainty to control loss weights and guide models to achieve satisfying results simultaneously, for instance, semantic segmentation and depth estimation. 
This technology is also well-developed in the depth estimation~\cite{kendall2017uncertainties,liu2019neural}, which significantly reduces noise of depth data. However, these depth estimation methods directly regressed the depth uncertainty by deep models and neglected specific geometry object-wise relationships, making them not quite suitable for the monocular 3D object detection topic. In this work, we try to compute uncertainty via combining both end-to-end learning and geometry relationships.

\subsection{Monocular 3D object detection} 
Monocular 3D object detection is a task of predicting 3D bounding boxes from a single image~\cite{ding2020learning,he2019mono3d++,kundu20183d,liu2019deep,manhardt2019roi,simonelli2019disentangling}, which includes object's 3D dimensions, yaw angle, and center coordinate. However, due to the inherent difficulties of this task, existing methods have taken different approaches.

\noindent 
\textbf{Solutions for 3D dimension and angle regression.}
Early approaches in monocular 3D object detection primarily focused on regression of 3D size and angle, rather than the more challenging task of depth estimation. Deep3DBox~\cite{mousavian20173d}, the first work in monocular 3D detection, effectively addressed the key angle prediction problem by leveraging geometric priors. It proposed to regress an alpha relative yaw angle for objects, avoiding ill-posed property caused by camera view. DeepMANTA~\cite{chabot2017deep} and RoI-10D~\cite{manhardt2019roi } introduced 3D CAD models to learn shape-based knowledge for objects, which improved the accuracy of 3D dimension predictions. These methods designed effective regression heads for 3D dimension and angle regression, achieving satisfactory results and allowing subsequent works to focus on the core problem of depth estimation.

\noindent 
\textbf{Geometry-consistency data augmentation.} Due to the ill-posed property of monocular 3D detection, traditional data augmentation used in 2D object detection may distort the geometry relationship between 2D and 3D, leading to performance degradation.
GeoAug~\cite{lian2022exploring} systematically analyzed different kinds of widely used data augmentation and categorized them as either geometry-consistent or not. This provides a guideline for selecting appropriate data augmentation methods in monocular 3D object detection.

\noindent 
\textbf{Monocular 3D suitable operator\&model designs} focused on designing better operators or models to improve the model's ability to represent objects. GS3D~\cite{li2019gs3d} used ROI surface features to extract more accurate object representations. It first computes a 3D bounding box and then constructs object features by applying RoI-pooling~\cite{girshick2015fast} to visible aspects of 3D bounding boxes, resulting in more geometrically accurate object representations. M3DRPN~\cite{brazil2019m3d} and D$^4$LCN~\cite{ding2020learning} proposed similar ideas, using region-specific convolutional kernels to create adaptive receptive fields that follow the "Closer appears larger and farther appears smaller" principle, to account for variance in geometry relationships between different spatial regions in images for monocular 3D detection. DEVIANT~\cite{kumar2022deviant} provided a more theoretical solution by introducing geometry deep learning and utilizing group-invariant convolution to naturally incorporate geometry-guided receptive fields, achieving high performance. MonoATT~\cite{zhou2023monoatt} proposes a specially designed transformer to cluster and merge tokens following geometry scheme, which has achieved state-of-the-art results for monocular 3D object detection. 

\noindent 
\textbf{Pseudo-3D representation methods} focused on preprocessing 2D monocular images into 3D representation data, such as point clouds or binocular images. It not only makes models benefit from the effectiveness of 3D representations but also allows well-developed point cloud or multi-view 3D detector designs to be applied to monocular 3D detection. The first work to achieve this idea is AM3D~\cite{ma2019accurate}, which transformed image into object-aware pseudo-LiDAR points. These pseudo point clouds combined 3D geometry information with 2D image color cues, achieving strong results with the aid of a well-designed point cloud detector, FrustumPNet~\cite{qi2018frustum}. Concurrent works also explored similar approaches to establish pseudo-LiDAR. Weng~\emph{et al.}~\cite{weng2019monocular} and Wang~\emph{et al.}~\cite{wang2019pseudo} proposed to extract scene-level pseudo-LiDAR, which also achieved good performance. PatchNet~\cite{ma2020rethinking} argued that point cloud formulation is not a key factor and proposed an image-based LiDAR representation that outperformed existing pseudo-LiDAR methods. Simonelli~\emph{et al.}~\cite{simonelli2021we} reported an issue with existing pseudo-LiDAR methods lacking additional depth training information, and proposed guiding pseudo-LiDAR models to predict 3D confidence, which achieved state-of-the-art pseudo-LiDAR results. Additionally, Pseudo-Stereo~\cite{chen2022pseudo} utilized depth estimation to introduce pseudo-binocular images and achieved excellent results. 

\noindent 
\textbf{Models with additional depth supervision.}
Pseudo-3D methods achieved good results because they all utilized an extra depth model trained on extra depth data to acquire pseudo point clouds. Inspired by this, many approaches have attempted to introduce depth as dense supervision into the end-to-end model. CaDNN~\cite{reading2021categorical} utilized depth to transfer the feature-level 2D representation into Bird's Eye View (BEV) features and treated monocular 3D detection as an end-to-end BEV detection task. M3OD~\cite{li2022diversity} fused depth regression results with geometry-based depth estimation, which resulted in state-of-the-art performance. DID-M3D~\cite{peng2022did} disentangled the object depth as a combination of scene depth and an object-specific depth bias, introducing an effective usage of depth supervision for monocular 3D detection. Although these methods can achieve satisfactory results, they all require extra dense depth to train the model, which is difficult to obtain on a large-scale level. Currently, in the monocular 3D detection field, researchers are focusing on developing models that can rely solely on object-wise depth supervision. Due to the challenges involved, most approaches attempt to introduce a geometry inductive bias to help the model address the ill-posed depth estimation problem.

\noindent 
\textbf{Geometry-based object depth estimation methods.} The exploration of geometry prior to the object depth estimation is mainly around scene understanding and geometry perspective projection. Scene understanding aims to infer depth from the scene cues while the perspective projection tends to derive the depth by object-wise geometry prior~\cite{bao2020object,barabanau2019monocular,cai2020monocular,ku2019monocular}. 

For scene understanding: MonoPair~\cite{chen2020monopair} proposed a pair-wise relationship to improve the monocular 3D detection performance. It built object graphs to model the spatial relationships between different objects and inferred more accurate depth. Homography Loss~\cite{gu2022homography} explored the scene constraints into training. It guided the 3D relative locations of detected objects same to the ground-truth relationships and achieved more general gains combined with several methods. PGD~\cite{wang2022probabilistic} established an object-wise 3D object graph to propagate depth information across objects, leading to more accurate predicted depth values combined with a simple baseline FCOS3D~\cite{wang2021fcos3d}. MonoDETR~\cite{zhang2022monodetr} constructed a foreground depth map to help the model understand the scene object depth spatial distributions with the aid of the DETR-liked model structure~\cite{carion2020end}. MonoEF~\cite{zhou2021monocular,zhou2021monoef} and MoGDE~\cite{zhou2022mogde} proposed to model the camera extrinsic changes and provided a grounded depth carrying the camera information, making the model achieve extremely accurate performance.

For perspective projection: Ivan~\emph{et al.}~\cite{barabanau2019monocular} combined the keypoint method and the projection to do geometry reasoning.  
Decoupled3D~\cite{cai2020monocular} used lengths of bounding box edges to project and get the inferred depth. Bao~\emph{et al.}~\cite{bao2020object} combined the center voting with the perspective projection to achieve better 3D center reasoning. All of these projection-based mono3D methods did not consider the error amplification problem, leading to limited performance. GUPNet~\cite{lu2021geometry} is the first work to tackle the error amplification problem in monocular 3D object detection, which utilizes a Geometry Uncertainty Projection module to compute reliable depth uncertainty. MonoRUn~\cite{chen2021monorun} investigated the uncertainty-driven PnP to infer the depth by dense constraints. MonoRCNN~\cite{shi2021geometry} and MonoRCNN++~\cite{shi2023multivariate} also established the geometry-based uncertainty for projection and achieved accurate results with the projection inductive bias. In this paper, we go further to solve the error amplification by uncertainty-based deep learning. 


\begin{figure*}[t]
\begin{center}
\includegraphics[width=0.8\linewidth]{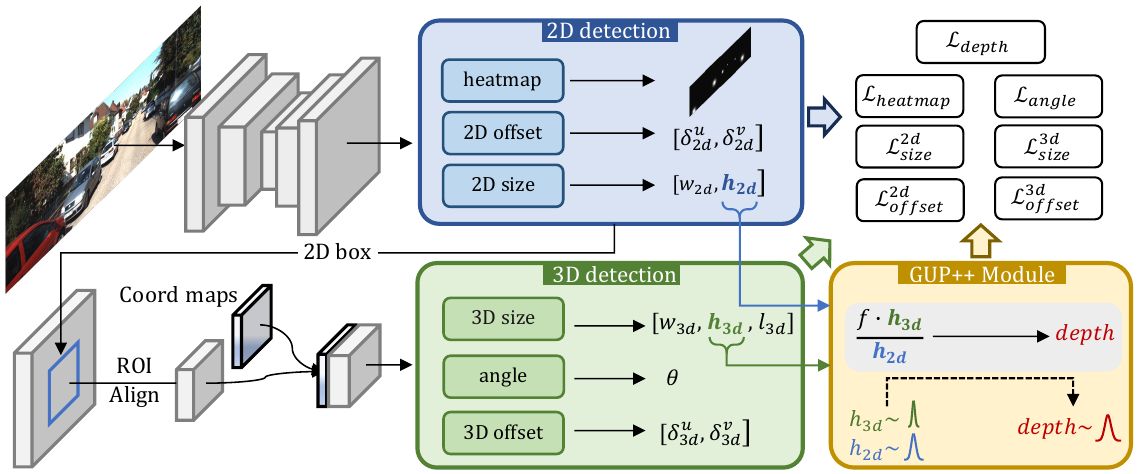}
\end{center}
   \caption{The framework of the GUPNet++. The input image is processed by the network to extract the 2D box and basic 3D box parameters. The Geometry Uncertainty Propagation module estimates the depth using the height parameters, helping both training and inference. } 
\label{fig:pipeline}
\end{figure*}
\section{Model Architecture}
Figure~\ref{fig:pipeline} illustrates the pipeline of the proposed Geometry Uncertainty Propagation Network (GUPNet++). The network takes an image as input and initially generates 2D bounding boxes using a 2D detection module. Subsequently, these 2D bounding boxes serve as the Regions of Interest (RoIs) during the 3D detection phase. For each RoI, our model predicts the items of angle, dimension, projected center, and depth of the corresponding 3D bounding box. 

\subsection{2D Detection}
We adopt the CenterNet \cite{zhou2019objects} with three heads to generate location, size, and confidence for each potential 2D bounding box.  As shown in Figure~\ref{fig:pipeline}, the backbone first extracts the feature maps with the size of $W\times H\times C$ to represent the input image, where $H$, $W$ and $C$ are height, width and channel number of the feature maps. 
Then, the heatmap head outputs the probability of category for each grid location. 
A 2D offset head computes bias $(\delta_{2d}^u,\delta_{2d}^v)$ to refine coarse location to accurate bounding box center, and a 2D size head obtains size $(w_{2d},h_{2d})$ for each box. 
%
These predicted 2D regions of interest (RoIs) are used in the subsequent 3D detection phase.

\noindent
\subsection{RoI Feature Representation}
Inspired by Mask R-CNN~\cite{he2017mask}, we employ RoIAlign to crop the RoI features and resize them to a fixed resolution of $7\times7\times C$. 
Then, we compute additional coordinate and category maps to enhance the RoI feature.

To compute the coordinate map, for each feature grid in the $7\times7$ RoI, we take its 2D coordinate $(u,v)$ in the \emph{original image} and then re-project it back to the 3D space by assuming its depth as 1 as follows:
\begin{equation}
\begin{aligned}
    &x^{np} = (u-c_u)/f,\\
    &y^{np} = (v-c_v)/f,\\
    &z^{np} = 1,
\end{aligned}    
\label{eq:cm}
\end{equation}
where $c_u$, $c_v$ and $f$ are the camera intrinsic parameters. $(x^{np},y^{np})$ at each grid are combined and result in a $7\times7\times2$ coordinate map. This coordinate map is concatenated with RoI feature, compensating for the missing absolute position and size cues caused by RoI cropping. The empirical studies show the significance of these cues~\cite{dijk2019neural}. Also, further introduced camera intrinsic parameters enable our model to account for geometry relationship variations when adapting to different cameras, making our model camera-independent.


For the category map, we broadcast 2D classification score, confidences for all $N$ (3 for KITTI~\cite{kitti} and 10 for nuScenes~\cite{caesar2020nuscenes} benchmarks.) categories, of the current RoI into the shape of $7\times7\times N$ and concatenate that maps also at the channel-wise level with RoI features. This additional category information has shown its effectiveness in the monocular 3D reconstruction task~\cite{tatarchenko2019single}, which would provide object prior to benefit the monocular 3D object detection. 

After that, each RoI feature is organized as a feature map with the shape $7 \times 7 \times (C+2+N)$.
%

\subsection{3D Detection}
Based on the RoI features, we construct several heads to predict 3D bounding boxes. In particular, similar to the 2D offset head, the 3D offset head estimates the bias $(\delta_{3d}^u,\delta_{3d}^v)$ between 3D center in image plane and coarse locations. The angle prediction head predicts the relative alpha rotation angle $\theta$~\cite{mousavian20173d} to indicate the direction of each object and the 3D size branch estimates the 3D dimension parameters, including height, width, and length, denoted as $(h_{3d},w_{3d},l_{3d})$. 
Finally, we design a probabilistic perspective projection module, called Geometry Uncertainty Propagation (GUP++) module, for depth estimation, which will be depicted in Section \ref{sec:gup++}.



\subsection{Loss Functions}
As introduced above, the proposed model has seven heads, including 2D heatmap, 2D offset, 2D size, yaw angle, 3D size, 3D offset, and depth. 

The heatmap is trained as a classification task by the Gaussian Focal Loss~\cite{lin2017focal,law2018cornernet}:
\begin{equation}
\begin{aligned}
\mathcal{L}_{heatmap} &= \frac{1}{N_{pos}}\sum_{uvc} \text{Focal}(P_{uvc},\hat{P}_{uvc}),
\end{aligned}
\end{equation}
where $P$ and $\hat{P}$ are the predicted and the ground-truth heatmaps, respectively. $u$, $v$, and $c$ means the spatial and channel index. $\text{Focal}(\bullet,\bullet)$ are defined as:
\begin{equation}
\begin{aligned}
\text{Focal}(y,\hat{y})&=\begin{cases}
-(1-y)^\alpha\cdot\text{log}(y), &\text{if}\ \hat{y}=1\\
(1-\hat{y})^\beta y^\alpha\text{log}(1-y)
,&\text{otherwise.}
\end{cases}
\end{aligned}
\end{equation}
Following~\cite{law2018cornernet}, a Gaussian kernel is applied for each positive location in $P^*$. $\alpha$ and $\beta$ are the hyper-parameters while $N_{pos}$ is the number of positive targets. 

For the yaw angle head, we use the \emph{MultiBin} loss.
The 360$^{\circ}$ angle range is equally split into 12 bins, and the angle classification term $\mathcal{L}_{conf}$ aims to guide the head to predict the corresponding bin category using the cross-entropy loss, whereas the angle residue regression term $\mathcal{L}_{res}$ guides the model to regress the gap between the coarse-grained bin and the accurate angle value using the L1 loss function:
\begin{equation}
\begin{aligned}
\mathcal{L}_{angle} = \mathcal{L}_{conf}(\theta,\hat{\theta})+\mathcal{L}_{res}(\theta,\hat{\theta}),
\end{aligned}
\end{equation}

For the two offset loss terms, we utilize the L1 regression loss as the following equations:
\begin{equation}
\begin{aligned}
\mathcal{L}_{offset}^{2d} = |\delta^u_{2d}-\hat{\delta}^{u}_{2d}|+|\delta^v_{2d}-\hat{\delta}^{v}_{2d}|,\\
\mathcal{L}_{offset}^{3d} = |\delta^u_{3d}-\hat{\delta}^{u}_{3d}|+|\delta^v_{3d}-\hat{\delta}^{v}_{3d}|,
\end{aligned}
\end{equation}
where $\hat{\delta}_{\bullet}$ means the ground-truth offsets. Similarly, the two size regression losses are defined as follows:
\begin{equation}
\begin{aligned}
\mathcal{L}_{size}^{2d} = \mathcal{L}_{h_{2d}}&+|w_{2d}-\hat{w}_{2d}|,\\
\mathcal{L}_{size}^{3d} = \mathcal{L}_{h_{3d}}+|w_{3d}&-\hat{w}_{3d}|+|l_{3d}-\hat{l}_{3d}|.
\end{aligned}
\end{equation}
We do not provide the formulas for heights and depth losses $\mathcal{L}_{h_{2d}}$, $\mathcal{L}_{h_{3d}}$ and $\mathcal{L}_{depth}$ here because their formulas are relevant to the requirements of the GUP++ module, which will be introduced in Section \ref{sec:gup++}.

\section{Geometry Uncertainty Propagation}
\label{sec:gup++}
Given the difficulty of directly regressing reliable depth from single images, we propose to leverage perspective projection to estimate depth by the projection process:
\begin{equation}
\label{eq:gep}
\begin{aligned}
    d_p=\frac{f\cdot h_{3d}}{h_{2d}}.
\end{aligned}
\end{equation}
where $d_p$, $f$, $h_{2d}$ and $h_{3d}$ are projected depth, camera focal length, object visual height and object 3D height, respectively. But unlike traditional methods providing single depth values, we formulate the projected depth within a probability manner. 

\subsection{Perspective Projection with Uncertainty Propagation}
\label{sec:geu_v2}
The GUP++ module first assumes 2D and 3D heights as distributions and the projection process is represented as:
\begin{equation}
\begin{aligned}
    D_p=\frac{f\cdot H_{3d}}{H_{2d}},
\end{aligned}
\end{equation}
where $D_p$, $H_{3d}$ and $H_{2d}$ mean projected depth, 3D height and 2D height random variables, respectively. 
Then, we assume $H_{2d}$ and $H_{3d}$ both follow Laplacian distributions $La(\mu_{2d}, \sigma_{2d})$ and $La(\mu_{3d}, \sigma_{3d})$, respectively. Their mean $\mu$ and standard deviation (std) $\sigma$ are predicted by corresponding 2D or 3D size regression heads. 
Currently, $D_p$ becomes a random variable with parameters of $(\mu_p,\sigma_p)$. The mean $\mu_{p}$ is obtained as:
\begin{equation}
\begin{aligned}
    \mu_{p} = \frac{f\cdot \mu_{3d}}{\mu_{2d}}.
\end{aligned}
\end{equation}
And the standard deviation $\sigma_{p}$ is computed by the uncertainty propagation~\cite{wiki:Propagation_of_uncertainty}:
\begin{equation}
\label{eq:geu++}
\begin{aligned}
    \sigma_{p} = &\sqrt{(\frac{\partial \mu_p}{\partial \mu_{2d}}\cdot\sigma_{2d})^2+(\frac{\partial \mu_p}{\partial \mu_{3d}}\cdot\sigma_{3d})^2}\\
    = &\mu_{p}\cdot\sqrt{\frac{\sigma^2_{2d}}{\mu^2_{2d}}+\frac{\sigma^2_{3d}}{\mu^2_{3d}}}.
\end{aligned}
\end{equation}
To obtain a better predicted depth, we add another learned stream to compute a bias to modify the initial projection results. The bias is also a distribution with the parameters of $(\mu_{b},\sigma_{b})$ and predicted by the model directly. Accordingly, the final depth distribution parameters are written as:
\begin{equation}
\label{eq:bias}
\begin{aligned}
     \mu_d = \mu_{p}+\mu_{b},\quad \sigma_{d} &= \sqrt{(\sigma_{p})^2+(\sigma_{b})^2}.
\end{aligned}
\end{equation}
We refer to the $\sigma_{d}$ as Geometry-guided Uncertainty plus (GeU++). 
Compared with the geometry-guided uncertainty (GeU) in the GUPNet~\cite{lu2021geometry}, the GeU++ reflects uncertainty from both $h_{2d}$ and $h_{3d}$ rather than single $h_{3d}$ . 
The consideration of 2D height makes the GUP++ module can directly train with the uncertainty-based loss function without the requirement of a heuristic scheme like Hierarchical Task Learning (HTL)~\cite{lu2021geometry}, which will be further described in Section~\ref{sec:un_e2e}.

Also, the GeU++ models the uncertainty propagated from heights to depth, making itself accurately estimate the error of the projected depth. With this, we can use this accurate uncertainty to measure the quality of depth projection and provide a score to indicate confidence for 3D bounding boxes, leading to highly reliable 3D detection, which will be further described in Section~\ref{sec:IAUS}.

%


\subsection{Training with Uncertainty-based Optimization}
\label{sec:un_e2e}
During training, we train 2D height, 3D height and depth by an improved Laplacian uncertainty loss function as follows:
\begin{equation}
\label{eq:beta-nll}
\begin{aligned}
     \mathcal{L}_{h_{2d}}=\lfloor\frac{\sigma_{2d}}{\sqrt{2}}\rfloor^\beta\left(\frac{\sqrt{2}}{\sigma_{2d}}|\mu_{2d}-\hat{h}_{2d}|+\log\sigma_{2d}\right),\\
     \mathcal{L}_{h_{3d}}=\lfloor\frac{\sigma_{3d}}{\sqrt{2}}\rfloor^\beta\left(\frac{\sqrt{2}}{\sigma_{3d}}|\mu_{3d}-\hat{h}_{3d}|+\log\sigma_{3d}\right),\\
     \mathcal{L}_{depth}=\lfloor\frac{\sigma_{d}}{\sqrt{2}}\rfloor^\beta\left(\frac{\sqrt{2}}{\sigma_d}|\mu_d-\hat{d}|+\log\sigma_d\right),\\
\end{aligned}
\end{equation}
where the $\lfloor\cdot\rfloor$ means the stop-gradient operation which blocks the gradient in the backward propagation. These losses are called $\beta$-Negative Log-Likelihood ($\beta$-NLL) loss. Note that, in the depth loss, we also assume the depth distribution follows the Laplace distribution for simplification. Next, we achieve the final loss by directly adding them together:
\begin{equation}
\begin{aligned}
     \mathcal{L}_{total} = \sum_{i\in \mathcal{T}} \mathcal{L}_{i},
\end{aligned}
\end{equation}
where the definition of $\mathcal{T}$ is the task set including heatmap, 2D size, 2D offset, 3D size, 3D offset, angle and depth.

Compared with the original GUPNet~\cite{lu2021geometry}, this new training procedure has several advantages: 1). We modify the original uncertainty loss function as a new one called the $\beta$-NLL Laplacian uncertainty loss, which makes the uncertainty training more effective. 2). The overall loss is computed by directly combining each task loss together without the need for the HTL~\cite{lu2021geometry} or any other curriculum learning scheme.

The original uncertainty loss, with the formulation shown as follows:
\begin{equation}
\label{eq:un_loss_general}
\begin{aligned}
\mathcal{L} =
    \frac{\sqrt{2}}{\sigma}|\mu-gt|+\log \sigma,
\end{aligned}
\end{equation}
which has a potentially negative effect as it reduces the overall loss weights when dealing with datasets with higher variance or more difficulty, such as nuScenes~\cite{caesar2020nuscenes}. Specifically, when the majority of samples have high uncertainty (large $\sigma$), the coefficient $\sqrt{2}/\sigma$ before the L1 term will be small, making most samples have low training weights. To address this issue, inspired by the $\beta$-NLL Gaussian loss~\cite{seitzer2022pitfalls},  we propose a modified $\beta$-NLL Laplacian uncertainty loss. 
In that loss, the hyper-parameter $\beta$, between 0$\sim$1, controls the trade-off between uncertainty-based learning and traditional L1 regression optimization. 
We follow the official $\beta$-NLL setting and set $\beta$ as 0.5~\cite{seitzer2022pitfalls}.

Based on the aforementioned loss, currently, 2D height,
3D height and depth estimations are all trained by the uncertainty
loss functions effectively. All of their training procedures are controlled by their corresponding uncertainties. And because of the \emph{uncertainty propagation mechanism} in the GUP++ module, the depth uncertainty contains both propagated uncertainties from 2D height and 3D height. So, where a sample has inaccurate 2D or 3D height estimations, it would correspondingly exhibit high depth uncertainty. Consequently, in the uncertainty-based loss, its depth training weight is small and only increases when the quality of the height estimations is improved. This mechanism allows depth will naturally adapt its loss weight after the two heights are tuned well. So the GUP++ can stable training for the perspective projection process, without the need for additional human-designed strategy. This property makes the GUP++ module be directly trained with the uncertainty loss function in an end-to-end manner, leading to better flexibility, such as combining with other detectors in the future. 

\begin{figure}[t]
\begin{center}
\includegraphics[width=0.9\linewidth]{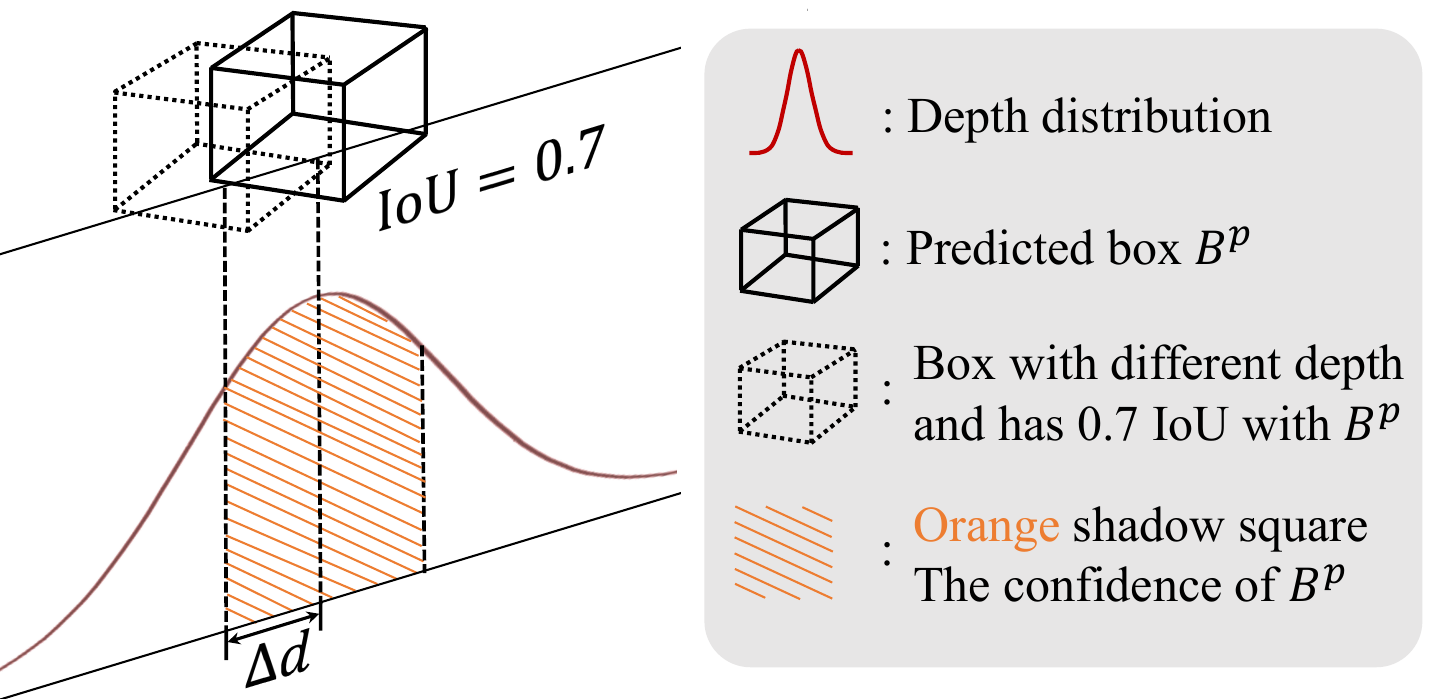}
\end{center}
   \caption{The computation pipeline of the IoU-guided Uncertainty-Confidence: The dashed line box means the furthest potential true box that has a 0.7 IoU value with our predicted box (the solid line one). Under that, the square of the \textcolor[RGB]{255 165 0}{orange} region under the depth distribution curve means the confidence of the predicted box $B^P$. }
\label{fig:box_dis}
\end{figure}

\subsection{Inference with IoU-guided Uncertainty-Confidence}
\label{sec:IAUS}
During inference, we set the depth mean $\mu_d$ and 3D height mean $\mu_{3d}$ as the depth and 3D height value of the predicted box $B^p$:
\begin{equation}
\label{eq:pred_box}
\begin{aligned}B^p=[\underbrace{\delta_{3d}^u,\delta_{3d}^v}_{\text{3d offset}},\underbrace{d[\mu_d]}_{\text{depth}},\underbrace{h_{3d}[\mu_{3d}],w_{3d},l_{3d}}_{\text{3d size}},\underbrace{\theta}_{yaw}],\\
\end{aligned}    
\end{equation}
whose confidence is written as $p_{3d}$ and derive as follow:
\begin{equation}
\label{eq:p3d}
\begin{aligned}
    p_{3d} = p_{3d|2d}\cdot p_{2d}.
\end{aligned}    
\end{equation}
This score combines both 2D detection confidence and 3D conditioned detection confidence, where $p_{2d}$ is extracted from the 2D detection. For $p_{3d|2d}$,
to make $p_{3d}$ more reliable, we propose a new IoU-guided Uncertainty Confidence (IoUnC) to derive it by the depth uncertainty $\sigma_d$ as follows:
\begin{equation}
\label{eq:iaus}
\begin{aligned}
p_{3d|2d}=1-exp(-\frac{\sqrt{2}\cdot\Delta d}{\sigma_d}),
\end{aligned}
\end{equation}
which means computing the cumulative probability in an interval of $[\mu_d-\Delta d,\mu_d+\Delta d]$:
\begin{equation}
\label{eq:cdf}
\begin{aligned}
p_{3d|2d}=\int_{\mu_d-\Delta d}^{\mu_d+\Delta d} p(D=x|\mu_d,\sigma_d) dx,
\end{aligned}
\end{equation}
where $p(D|\mu_d,\sigma_d)$ is the depth distribution assumed as a Laplacian distribution before. $\Delta d$ is computed by:
\begin{equation}
\label{eq:delta_d}
\begin{aligned} \Delta d = max\{d' | \text{IoU}(B^p(d=\mu_d+d'),B^p) \geq th\}.
\end{aligned}
\end{equation}
$\text{IoU}(A,B)$ means computing the Intersection over Union (IoU) value between boxes $A$ and $B$. $B^p(d=x)$ means changing the depth of $B^p$ manually as a given value $x$. $th$ is a hyper-parameter which is set as 0.7 in this work. The computation pipeline of IoUnC is organized as shown in Figure~\ref{fig:box_dis}. 

This new score can be understood as measuring the width of the depth distribution. For a high uncertainty/variance, the depth distribution is wide and leads to low score. The accumulation interval parameter $\Delta d$ is determined by object properties, dimension and orientation. Such a design addresses the characteristics of different objects. In the case of an object with a large size, $\Delta d$ should be large. This aligns with our intuitive understanding that we are more tolerant of larger error for large objects~\footnote{This property is reflected in object detection evaluation metric Intersection over Union (IoU). For example, during inference, for the same-level depth estimation error, the big-size object will have a larger IoU value with its target box and the small-size object will not. It implies that the depth tolerance is object-specific, which is related to object size and orientation}. This makes the IoUnC provide more flexible confidence and leads to better detection results. With such IoUnC, the whole model can make a reliable output because the confidence computed by this scheme can reflect the uncertainty from the height estimation to the depth which makes such confidence really reflect the quality of the predicted box.

\begin{table*}[!ht]
\centering
\fontsize{8}{10}\selectfont
\caption{{\bf 3D object detection on the KITTI \emph{test} set.} We highlight the best results in {\bf bold} and the second results in \underline{underline}. For the extra data: 1). `Depth' means the methods use extra depth annotations or off-the-shelf networks pre-trained from a larger depth estimation dataset. 2). `Temporal' means using additional temporal data. 3). `LiDAR' means utilizing real LiDAR data for better training. 4) 'Camera Extrinsic Pose' means utilizing camera intrinsic pose supervision or model like DeepVP~\cite{fang2022deepvp}. 5). `None' denotes no extra data is used.}
\label{tab:kitti_test}
\begin{tabular}{l||c|ccc|ccc|ccc}
\toprule
\multirow{2}{*}{Method}&Extra data/&
\multicolumn{3}{c|}{Car@IoU=0.7}&\multicolumn{3}{c|}{Pedestrian@IoU=0.5}&
\multicolumn{3}{c}{Cyclist@IoU=0.5} \cr\cline{3-11} &modules&
Easy&Mod.&Hard&Easy&Mod.&Hard&Easy&Mod.&Hard\cr\hline
Decoupled-3D~\cite{cai2020monocular}&Depth
& 11.08 & 7.02 & 5.63
& – & – & –
& – & – & –\cr
Mono-PLiDAR~\cite{weng2019monocular}&Depth
& 10.76 & 7.50 & 6.10
& – & – &–
& – & – &–  \cr
AM3D~\cite{ma2019accurate}&Depth
& 16.50 & 10.74 & 9.52 
& – & – & – 
& – & – & –\cr
PatchNet~\cite{ma2020rethinking}&Depth
& 15.68 & 11.12 & 10.17
& – & – & –
& – & – & –  \cr
DA-3Ddet~\cite{da3dnet}&Depth
& 16.77 & 11.50 & 8.93
& – & – & –
& – & – & –\cr
D4LCN~\cite{ding2020learning}&Depth
& 16.65 & 11.72 & 9.51
& 4.55 & 3.42 & 2.83
& 2.45 & 1.67 & 1.36\cr
DID-M3D~\cite{peng2022did}&Depth
&21.99&15.39&12.73
& – & – & –
& – & – & –\cr
MonoDTR~\cite{huang2022monodtr}&Depth
&24.40&16.29&13.75
&15.33&10.18&8.61
&5.05&3.27&3.19\cr
DD3D~\cite{peng2022did}&Depth
&23.19&16.87&14.36
&16.64&11.04&9.38
&7.52&4.79&4.22\cr
CMKD~\cite{hong2022cross}&Depth
&25.09&16.99&15.30
&17.79&11.69&10.09 
&9.60&5.24&4.50 \cr
MonoDDE~\cite{li2022diversity}&Depth
&24.93&17.14&14.10
&11.13&7.32&6.67 
&5.84&3.78&3.33\cr
MonoPGC~\cite{wu2023monopgc}&Depth
&24.68&17.17&14.14
&14.16&9.67&8.26
&5.88&3.30&2.85\cr
Kinematic~\cite{brazil2020kinematic}&Temporal
& 19.07 & 12.72 & 9.17
& – & – & –
& – & – & –\cr
MonoPSR~\cite{ku2019monocular}&LiDAR
& 10.76 & 7.25 & 5.85
& 6.12 & 4.00 & 3.30
& 8.70 & 4.74 & 3.68\cr
CaDNN~\cite{reading2021categorical}&LiDAR
& 19.17 & 13.41 & 11.46
& 12.87 & 8.14 & 6.76
& 7.00 & 3.41 & 3.30 \cr
MonoDistill~\cite{chong2022monodistill}&LiDAR
&22.97&16.03&13.60
& – & – & –
& – & – & –\cr
Autoshape~\cite{liu2021autoshape}&CAD
&22.47&14.17&11.36
& – & – & –
& – & – & –\cr
MonoEF~\cite{zhou2021monocular}&Camera Extrinsic Pose
&21.29&13.87&11.71
&4.27&2.79&2.21
&1.80&0.92&0.71\cr
MoGDE~\cite{zhou2022mogde}&Camera Extrinsic Pose
&27.07&17.88&15.66
& – & – & –
& – & – & –\cr
\hline
MonoDIS~\cite{simonelli2019disentangling}&None
& 10.37 & 7.94 & 6.40
& – & – & –
& – & – & –\cr
UR3D~\cite{ur3d}&None
& 15.58 & 8.61 & 6.00
& – & – & –
& – & – & –\cr
M3D-RPN~\cite{brazil2019m3d}&None
& 14.76 & 9.71 & 7.42
& 4.92 & 3.48 & 2.94
& 0.94 & 0.65 & 0.47\cr
SMOKE~\cite{smoke}&None
& 14.03 & 9.76 & 7.84
& – & – & –
& – & – & –\cr
MonoPair~\cite{chen2020monopair}&None
& 13.04 & 9.99 & 8.65
& 10.02 & 6.68 & 5.53
& 3.79 & 2.12 & 1.83\cr
RTM3D~\cite{li2020rtm3d}&None
& 14.41 & 10.34 & 8.77
& – & – & –
& – & – & –\cr
MoVi-3D~\cite{movi3d}&None
& 15.19 & 10.90 & 9.26
& 8.99 & 5.44 & 4.57
& 1.08 & 0.63 & 0.70\cr
ImVoxelNet~\cite{rukhovich2022imvoxelnet}&None
&17.15&10.97&9.15
&-&-&-
&-&-&-\cr
RAR-Net~\cite{rarnet}&None
& 16.37 & 11.01 & 9.52
& – & – & –
& – & – & –\cr
PGD~\cite{wang2022probabilistic}&None
&19.05&11.76&9.39
&2.28&1.49&1.38
&2.81&1.38&1.20\cr
ImVoxelNet~\cite{rukhovich2022imvoxelnet}+Homoloss~\cite{gu2022homography}&None
&20.10&12.99&10.50
&\underline{12.47}&7.62&6.72
&1.52&0.85&0.94\cr
MonoFlex~\cite{zhang2021objects}&None
&19.94&13.89&12.07
&9.43&6.31&5.26
&4.17&2.35&2.04\cr
MonoFlex~\cite{zhang2021objects}+Homoloss~\cite{gu2022homography}&None
&21.75&14.94&13.07
&11.87&7.66&6.82
&\underline{5.48}&\underline{3.50}&\underline{2.99}\cr
DCD~\cite{li2022densely}&None
&23.81&15.90&13.21
&10.37&6.73&6.28
& 4.72&2.74&2.41\cr
MonoDETR~\cite{zhang2022monodetr}&None&\underline{24.52}&\underline{16.26}&\underline{13.93}& – & – & – & – & – & – \cr
\hline
GUPNet~\cite{lu2021geometry} &None
&20.11&14.20&11.77
&\bf{14.72}&\bf{9.53}&\bf{7.87}
& 4.18 & 2.65 & 2.09 \cr
GUPNet++ (Ours) &None
&\bf{24.99}&\bf{16.48}&\bf{14.58}
&12.45&\underline{8.13}&\underline{6.91}
&\bf{6.71}&\bf{3.91}&\bf{3.80}\cr
\bottomrule  
\end{tabular}
\end{table*}

\section{Experiments}
\subsection{Setup}
\noindent
\textbf{Dataset.} We evaluate our method on KITTI and nuScenes benchmarks. The KITTI 3D dataset~\cite{kitti} is the most commonly used benchmark in 3D object detection, and it provides left-camera images,  calibration files, and annotations for standard monocular 3D detection. 
It totally provides 7,481 frames for training and 7,518 frames for testing.
Following~\cite{mono3d,mv3d}, we split the training data into a training set (3,712 images) and a validation set (3,769 images). 
We conduct ablation studies based on this split and also report the final results with the model trained on all 7,481 images and tested by KITTI official server. The nuScenes dataset~\cite{caesar2020nuscenes} is a comprehensive dataset that contains multi-modal data from 1000 scenes, including RGB images from six surrounding cameras, points from five radars, and one LiDAR. The dataset is split into 700 scenes for training, 150 scenes for validation, and 150 scenes for testing. There are a total of 1.4 million annotated 3D bounding boxes from ten categories. We also report our method on this large-scale dataset to show its generalization.

\noindent
\textbf{Evaluation protocols.} For KITTI, all the experiments follow the standard evaluation protocol in the monocular 3D object detection and bird's view (BEV) detection tasks. 
Following~\cite{simonelli2019disentangling}, we evaluate the model by metrics of ${\rm AP}_{40}$ and ${\rm AP}_{11}$. And for nuScenes, we follow the commonly used metrics to evaluate the performance, including mAP, Average
Translation Error (ATE), Average Scale Error (ASE), Average Orientation Error (AOE), mean TP metric (mTP) and nuScenes Detection Score (NDS). Please note that we do not provide results about Average Velocity Error (AVE) and Average Attribute Error (AAE) because these two metrics are not suitable for our methods here because we follow the setting of CenterNet~\cite{zhou2019objects,zhou2020tracking} that does not predict attribute and velocity. 

\noindent
\textbf{Implementation details.} For both dataset, we use DLA-34~\cite{yu2018deep} as our backbone for both baseline and our method. Each 2D detection head has two convolutional (Conv) layers (the channel of the first one is set to 256) and each 3D head includes one 3x3 Conv layer with 256 channels, one averaged pooling layer and one fully-connected layer. The output channels of these heads depend on the output data structure. The whole model is totally trained for 140 epochs and the initial learning rate is 1.25$e^{-3}$, which is decayed by 0.1 at the 90-th and the 120-th epoch. To make the training more stable, we apply the linear warm-up strategy in the first 5 epochs. Further, for KITTI, the resolution of the input image is set to 380 × 1280 and the batch size is set as 32 which is the same as the GUPNet~\cite{lu2021geometry}. And for nuScenes, the batch size and the image resolution are modified to 128 and 448 × 800, both following the popular DLA series detector CenterNet~\cite{zhou2019objects,zhou2020tracking}.

To enhance the GUPNet++ further, we implement following enhancements for the model: 1). We modify Non-maximum Suppression (NMS) scheme from 2D level to 3D level. Specifically, the NMS is done for 3D bounding boxes rather than 2D boxes. 2). Following the approach of CenterNet~\cite{zhou2019objects,zhou2020tracking}, we incorporate the Deformable Convolution Network (DCN)~\cite{dai2017deformable} layer into the DLA backbone, making the receptive field of the backbone network more flexible.
Other settings are the same as the original GUPNet. Also, for the nuScenes benchmark, we provide additional results on a large-scale backbone HGLS-104~\cite{newell2016stacked} network. 
\subsection{Main Results}

\begin{table*}[!t]
\begin{center}
\fontsize{8}{10}\selectfont
\caption{{\bf Performance of the Car category on the KITTI \emph{validation} set.} 
We highlight the best results in {\bf bold} and the second results in \underline{underline}.}
\label{tab:kitti_val}
\begin{tabular}{l||ccc|ccc|ccc|ccc}
\toprule
\multirow{2}{*}{Method} & \multicolumn{3}{c|}{3D@IoU=0.7} & \multicolumn{3}{c|}{BEV@IoU=0.7} & \multicolumn{3}{c|}{3D@IoU=0.5} & \multicolumn{3}{c}{BEV@IoU=0.5}\\ 
\cline{2-13} 
 ~ & Easy & Mod. & Hard  & Easy & Mod. & Hard & Easy & Mod. & Hard & Easy & Mod. & Hard \\ 
\hline
CenterNet~\cite{zhou2019objects} 
& 0.60 & 0.66 & 0.77 
& 3.46 & 3.31 & 3.21 
& 20.00 & 17.50 & 15.57
& 34.36 & 27.91 & 24.65 \\  
MonoGRNet~\cite{qin2019monogrnet} 
& 11.90 & 7.56  & 5.76 
& 19.72 & 12.81 & 10.15 
& 47.59 & 32.28 & 25.50
& 48.53 & 35.94 & 28.59 \\  
MonoDIS~\cite{simonelli2019disentangling}  
& 11.06 & 7.60 & 6.37 
& 18.45 & 12.58 & 10.66 
& - & - &
& - & - &\\  
M3D-RPN~\cite{brazil2019m3d}
& 14.53 & 11.07 & 8.65 
& 20.85 & 15.62 & 11.88 
& 48.53 & 35.94 & 28.59
& 53.35 & 39.60 & 31.76\\ 
MoVi-3D~\cite{movi3d}	
&14.28  &11.13  &9.68      
&22.36  &17.87  &15.73
&-		&-		&-		
&-		&-		&-	\\
MonoPair~\cite{chen2020monopair}
& 16.28 & 12.30 & 10.42
& 24.12 & 18.17 & 15.76
& 55.38 & {42.39} & {37.99}
& 61.06 & {47.63} & {41.92}\\ 
MonoDLE~\cite{ma2021delving}
& {17.45} & {13.66} & {11.68} 
& {-} & {-} & -
& {-} & - & -
& {-} & - & -\cr
MonoGeo~\cite{zhang2021learning}
& {18.45} & {14.48} & {12.87} 
& {-} & {-} & -
& {-} & - & -
& {-} & - & -\cr
PGD~\cite{wang2022probabilistic}
& {19.27} & {13.23} & {10.65} 
& {-} & {-} & -
& {-} & - & -
& {-} & - & -\cr
MonoFlex~\cite{zhang2021objects}
& {23.64} & {17.51} & {14.83} 
& {-} & {-} & -
& {-} & - & -
& {-} & - & -\cr
MoGDE~\cite{zhou2022mogde}
& {23.35} & {20.35} & \underline{17.71} 
& {-} & {-} & -
& {-} & - & -
& {-} & - & -\cr 
MonoDETR~\cite{zhang2022monodetr}
& \underline{28.84} & \bf{20.61} & {16.38} 
& \underline{37.86} & \underline{26.94} & \underline{22.79}
& \bf{68.85} & \underline{48.92} & \underline{43.57}
& \bf{72.30} & \underline{53.09} & \underline{46.61}\cr
\hline
GUPNet~\cite{lu2021geometry}
& {22.76} & {16.46} & {13.72} 
& {31.07} & {22.94} & {19.75} 
& {57.62} & 42.33 & 37.59 
& {61.78} & 47.06 & 40.88\cr
GUPNet++ (Ours)
& \bf{29.03} & \underline{20.45} & \bf{17.89} 
& \bf{38.82} & \bf{27.95} & \bf{24.96} 
& \underline{66.66} & \bf{49.65} & \bf{45.23}
& \underline{71.55} & \bf{54.00} & \bf{49.34}\cr 
\bottomrule
\end{tabular}
\end{center}
\end{table*}

\begin{table*}
\centering
\fontsize{8}{10}\selectfont
\caption{{\bf $\text{AP}_{11}$ results of the Car category on the KITTI \emph{validation} set.} 
We highlight the best results in {\bf bold}.}
\label{tab:kitti_val_ap11}
\begin{tabular}{l||ccc|ccc|ccc|ccc}
\toprule
\multirow{2}{*}{Method} & \multicolumn{3}{c|}{3D@IoU=0.7} & \multicolumn{3}{c|}{BEV@IoU=0.7} & \multicolumn{3}{c|}{3D@IoU=0.5} & \multicolumn{3}{c}{BEV@IoU=0.5}\\ 
\cline{2-13} 
 ~ & Easy & Mod. & Hard  & Easy & Mod. & Hard & Easy & Mod. & Hard & Easy & Mod. & Hard \\ 
\hline
Mono3D~\cite{chen2016monocular} 
& 2.53 & 2.31 & 2.31 
& 5.22 & 5.19 & 4.13
& - & - &
& - & - &\\
OFTNet~\cite{roddick2018orthographic} 
& 4.07 & 3.27 & 3.29 
& 11.06 & 8.79 & 8.91 
& - & - &
& - & - &\\
Deep3DBox~\cite{mousavian20173d} 
& 5.85 & 4.19 & 3.84
& 9.99 & 7.71 & 5.30 
& 27.04 & 20.55 & 15.88
& 30.02 & 23.77 & 18.83 \\  
FQNet~\cite{liu2019deep}
& 5.98 & 5.50 & 4.75
& 9.50 & 8.02 & 7.71
& 28.16 & 28.16 & 28.16
& 32.57 & 24.60 & 21.25\\
Mono3D++~\cite{he2019mono3d++} 
&10.60&7.90&5.70&16.70&11.50&10.10
&42.00&29.80&24.20
&46.70&34.30& 28.10\\
GS3D~\cite{li2019gs3d}
&13.46 & 10.97 & 10.38
& - & - &
&32.15  &29.89  &29.89		
& - & - & \\ 
MonoGRNet~\cite{qin2019monogrnet} 
& 13.88 & 10.19 & 7.62
& 24.97 & 19.44 & 16.30
& 50.51 & 36.97 & 30.82
& 54.21 & 39.69 & 33.06\\
MonoDIS~\cite{simonelli2019disentangling}  
& 18.05 & 14.98 & 13.42 
& 24.26 & 18.43 & 16.95
& - & - &
& - & - &\\  
M3D-RPN~\cite{brazil2019m3d}
& 20.27 & 17.06 & 15.21
& 25.94 & 21.18 & 21.18
& 48.96 & 39.57 & 33.01
& 53.35 & 39.60 & 31.76\\ 
RTM3D~\cite{li2020rtm3d}	
&20.77  &20.77  &16.63      
&25.56  &22.12  &20.91
&54.36  &41.90	&35.84		
&57.47	&44.16	&42.31\\ 
RARNet~\cite{rarnet}+GS3D~\cite{li2019gs3d}
&11.63 & 10.51 & 10.51
&14.34  &12.52  &11.36
&30.60  &26.40  &22.89		
&38.24  &32.01 &28.71 \\ 
RARNet~\cite{rarnet}+MonoGRNet~\cite{qin2019monogrnet}
&13.84 & 10.11 & 7.59
&24.84 & 19.27 & 16.20
&50.27 & 36.67 & 30.53
&53.91 & 39.45 & 32.84  \\ 
RARNet~\cite{rarnet}+M3D-RPN~\cite{brazil2019m3d}
&23.12  &19.82  &16.19
&29.16  &22.14  &18.78
&51.20  &44.12  &32.12		
&57.12  &44.41  &37.12 \\ \hline
GUPNet~\cite{lu2021geometry}   
& 25.76 & 20.48 & 17.24 
& 34.00 & 24.81 & 22.96 
& 59.36 & 45.03 & 38.14 
& 62.58 & 46.82 & 44.81\cr 
GUPNet++   
& {\bf 32.61} & {\bf 24.95} & {\bf 22.67}
& {\bf 41.23} & {\bf 31.50} & {\bf 29.16}
& {\bf 66.04} & {\bf 50.88} & {\bf 46.44}
& {\bf 70.37} & {\bf 54.78} & {\bf 50.95}\cr 
\bottomrule
\end{tabular}
\end{table*}

\begin{table*}[!t]
    \centering
    \fontsize{8}{10}\selectfont
    \caption{{\bf Ablation studies} on the KITTI \emph{validation} set for the Car category.}
    \label{tab:ablation}
    \begin{tabular}{l||cc|ccc|ccc|cc|ccc}
        \toprule
        \multirow{3}{*}{}&\multicolumn{5}{c|}{GUPNet}&\multicolumn{5}{c|}{GUPNet++}&\multicolumn{3}{c}{\multirow{2}{*}{3D@IoU=0.7}}\cr\cline{2-11} 
        &\multicolumn{2}{c|}{Base detector}&\multicolumn{3}{c|}{GUP}&\multicolumn{3}{c|}{GUP++}&\multicolumn{2}{c|}{Model enhancements}& & & \cr\cline{2-14} 
        &RoI-FR&GeP&UnC&GeU&HTL&GeU++&IoUnC&$\beta$-NLL&NMS3D&DCN&Easy&Mod.&Hard\cr\hline
        (a)&-&-&-&-&-&-&-&-&-&-&15.18&11.00&9.52\cr
        (b)&\checkmark&-&-&-&-&-&-&-&-&-&16.39&12.44&11.01\cr
        (c)&\checkmark&\checkmark&-&-&-&-&-&-&-&-&17.27&12.79&10.51\cr
        (d)&\checkmark&-&\checkmark&-&-&-&-&-&-&-&19.69&13.53&11.33\cr 
        (e)&\checkmark&\checkmark&\checkmark&-&-&-&-&-&-&-&18.23&13.57&11.22\cr
        (f)&\checkmark&\checkmark&\checkmark&\checkmark&-&-&-&-&-&-&20.86&15.70& 13.21\cr
        (g)&\checkmark&\checkmark&\checkmark&-&\checkmark&-&-&-&-&-&21.00&15.63& 12.98\cr 
        (h)&\checkmark&\checkmark&\checkmark&\checkmark&\checkmark&-&-&-&-&-&22.76&16.46&13.72\cr\hline
        (i)&\checkmark&\checkmark&\checkmark&-&\checkmark&\checkmark&-&-&-&-&23.90&16.71&14.40\cr
        (j)&\checkmark&\checkmark&\checkmark&-&-&\checkmark&-&-&-&-&23.56& 16.53&14.24\cr
        (k)&\checkmark&\checkmark&-&-&-&\checkmark&\checkmark&-&-&-&24.76& 17.43&14.78\cr
        (l)&\checkmark&\checkmark&-&-&-&\checkmark&\checkmark&\checkmark&-&-&26.08&17.85&15.10\cr\hline
        (m)&\checkmark&\checkmark&-&-&-&\checkmark&\checkmark&\checkmark&\checkmark&-&26.49&19.06&16.55\cr
        (n)&\checkmark&\checkmark&-&-&-&\checkmark&\checkmark&\checkmark&\checkmark&\checkmark&29.03&20.45&17.89\cr
        \bottomrule        
    \end{tabular}
\end{table*}

\begin{table*}[!ht]
\centering
\fontsize{8}{10}\selectfont
\caption{{\bf Performance on the nuScenes \emph{test} set.} We highlight the best results in {\bf bold} and the second results in \underline{underline}. }
\label{tab:nus_test}
\begin{tabular}{c||ccc|ccccccc}
\toprule
Method&Image Size&Backbone&Input Type&mAP↑&mATE↓&mASE↓&mAOE↓&NDS↑\cr\hline
PETR-R50~\cite{liu2022petr}&1056 × 384&ResNet-50~\cite{he2016deep}&Multi-view&31.3&0.768&0.278&0.564&0.381\cr
PETR-R101~\cite{liu2022petr}&1408 × 512&ResNet-101~\cite{he2016deep}&Multi-view&35.7&0.710&0.270&0.490&0.421\cr
BEVDet~\cite{huang2021bevdet}&704 × 256&SwinTransformer-Tiny~\cite{liu2021swin}&Multi-view&31.0&0.681&0.273&0.570&0.387\cr
BEVDet~\cite{huang2021bevdet}+DiffBEV~\cite{zou2023diffbev}&704 × 256&SwinTransformer-Tiny~\cite{liu2021swin}&Multi-view&31.5&0.660&0.265&0.567&0.398\cr\hline
FCOS3D~\cite{wang2021fcos3d}&1600 × 900&ResNet-101~\cite{he2016deep}&Monocular&29.5&0.806&0.268&0.511&0.372\cr
MonoDis~\cite{simonelli2019disentangling}&-&ResNet-34~\cite{he2016deep}&Monocular&30.4&0.738&0.263&0.546&0.384\cr
CenterNet-DLA34~\cite{zhou2019objects}&800 × 448&DLA-34~\cite{yu2018deep}&Monocular&30.6&0.716&0.264&0.609&0.328\cr
CenterNet-HGLS104~\cite{zhou2019objects}&-&HGLS-104~\cite{newell2016stacked}&Monocular&\underline{33.8}&0.658&0.255&0.629&0.400\cr
PGD~\cite{wang2022probabilistic}&1600 × 900&ResNet-101~\cite{he2016deep}&Monocular&33.5&0.732&0.263&\underline{0.423}&\bf{0.409}\cr\hline
GUPNet++ (ours)&800 × 448&DLA-34~\cite{yu2018deep}&Monocular&33.7&\underline{0.623}&\underline{0.251}&\bf{0.422}&0.399\cr
GUPNet++ (ours)&800 × 448&HGLS-104~\cite{newell2016stacked}&Monocular&\bf{34.8}&\bf{0.615}&\bf{0.246}&0.451&\underline{0.402}\cr
\bottomrule  
\end{tabular}
\end{table*}

\noindent
{\bf Results on the KITTI \emph{test} set.}
Table~\ref{tab:kitti_test} presents the results of our method compared to other approaches on the KITTI test set. Our proposed method demonstrates superior performance in the \emph{Car} category compared to previous methods.
Under fair conditions, without the use of extra data or modules, our method achieves comparable results to the state-of-the-art transformer-based method MonoDETR~\cite{zhang2022monodetr} and outperforms most convolutional neural network (CNN) methods by a significant margin. Notably, when compared to the CNN-based state-of-the-art method DCD~\cite{li2022densely}, GUPNet++ achieves improvements of approximately 1.18\%, 0.58\%, and 1.37\% on the Easy, Moderate, and Hard settings, respectively, demonstrating the effectiveness of the newly proposed GUPNet++.
Furthermore, we report the detection results for \emph{Pedestrian} and \emph{Cyclist} categories in Table~\ref{tab:kitti_test}. Our GUPNet and GUPNet++ models outperform all competing methods trained without extra data across all difficulty levels for pedestrian and cyclist detection. This highlights the generalization capability of our methods in adapting to multiple category detection scenarios.
Moreover, when compared to methods trained with extra data or modules, GUPNet++ achieves comparable results in most cases, further validating the effectiveness of our approach. 

\noindent
{\bf Results of Car category on the KITTI \emph{validation} set.}
To provide a comprehensive comparison, we present the performance of our model on the KITTI validation set in Table~\ref{tab:kitti_val}, considering various tasks and IoU thresholds.
At the 0.5 IoU threshold, our method achieves gains over the best competing method MonoDETR in several metrics. This demonstrates the effectiveness of our approach across different evaluation criteria.
Furthermore, our method consistently outperforms MonoDETR in 3D and BEV detection under most metrics at the 0.7 IoU threshold. This indicates that our method is particularly suitable for high-precision tasks, which is crucial in the context of autonomous driving.
Additionally, in Table~\ref{tab:kitti_val_ap11}, we compare our method with others using the ${\rm AP}{11}$ metric for methods that do not report the ${\rm AP}{40}$ metric on the validation set. This allows for a fair comparison among these methods.
Overall, the results on the KITTI validation set demonstrate the strong performance of our method, highlighting its effectiveness in various evaluation scenarios.

\noindent
{\bf Results of the nuScenes benchmark.}
We present the performance of our GUPNet++ on the large-scale nuScenes benchmark using two backbone settings: DLA-34 and HGLS-104. The results are summarized in Table~\ref{tab:nus_test}.

For the lightweight version with DLA-34 as the backbone, our method achieves comparable results to the state-of-the-art methods PGD (with a large-scale ResNet-101 backbone and graph post-processing module) and CenterNet-HGLS104. Our GUPNet++ achieves 33.7\% mAP, surpassing PGD and performing comparably with CenterNet-HGLS104. Additionally, our model achieves top results in terms of mATE, mASE and mAOE. Compared to CenterNet-DLA34, which shares the same backbone as our model, our GUPNet++ yields a significant improvement of approximately 3.2\% in mAP. Moreover, our model outperforms CenterNet-DLA34 by 0.093 in mATE, 0.013 in mASE, 0.187 in mAOE, and 0.071 in NDS, respectively, showcasing the effectiveness of our proposed GUPNet++ model.

Furthermore, we provide a large-scale version with HGLS-104 as the backbone, which achieves a new state-of-the-art performance for monocular 3D object detection on the nuScenes benchmark. It achieves 34.8\% mAP without the need for additional techniques such as test time augmentation, re-finetuning, or ensemble methods. Our model outperforms other models by a significant margin, demonstrating the generalization of our proposed GUPNet++ in combination with different backbones and reaffirming the effectiveness of our model.

\noindent
{\bf Latency analysis.}
We also test the running time of our system (DLA-34 backbone). We test the averaged running time on a single Nvidia TiTan XP GPU and achieve 29.4 FPS, which shows the efficiency of the inference pipeline. 

\begin{figure*}[t]
\begin{center}
\includegraphics[width=1.0\linewidth]{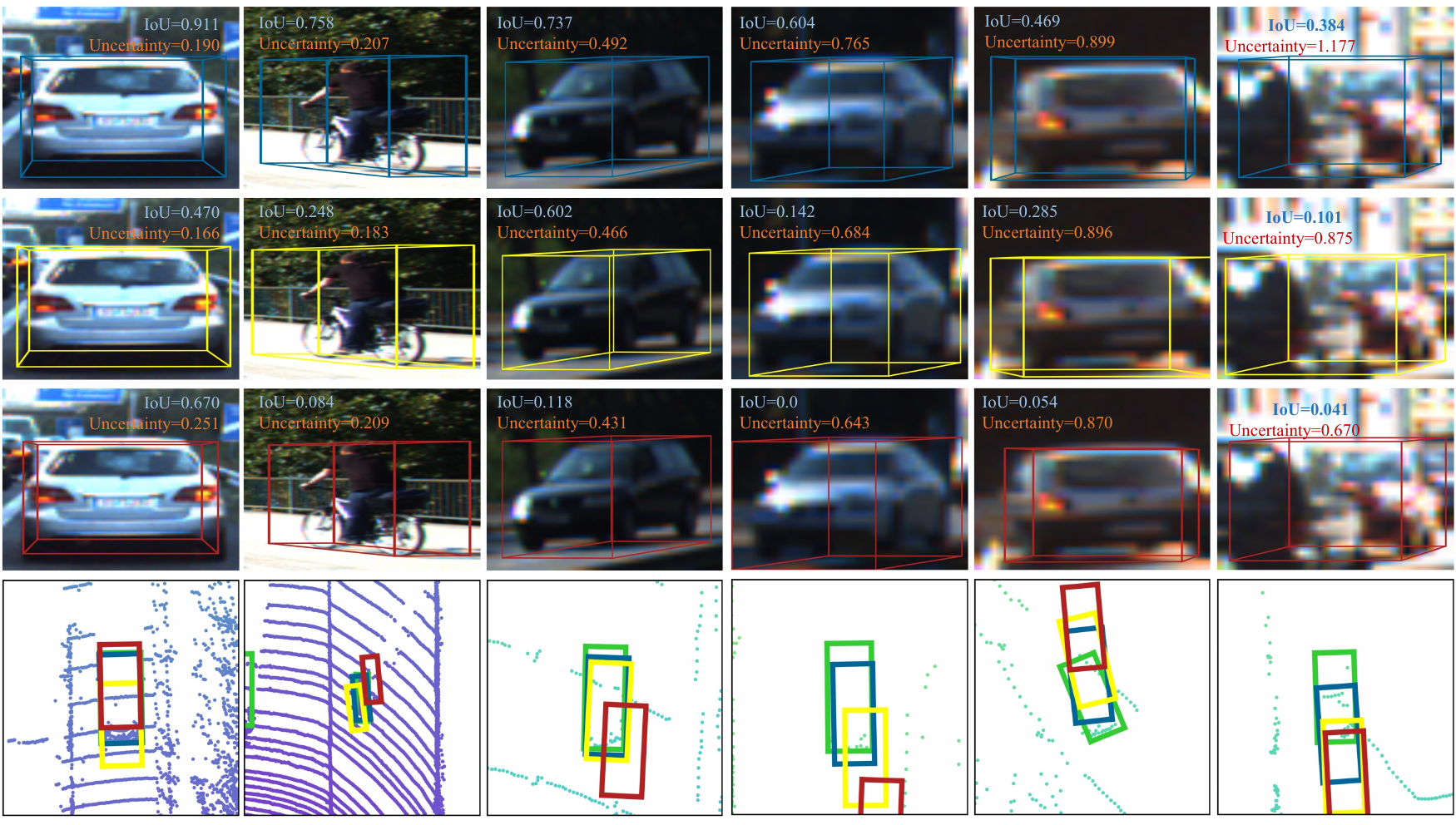}
\end{center}
   \caption{The visualized uncertainty examples on the validation set. The first row (\textcolor[RGB]{0,102,153}{Blue} boxes) are the results of our GUPNet++ while the second row (\textcolor[RGB]{235,235,35}{Yellow} boxes) are the GUPNet results. And the third row (\textcolor[RGB]{178,34,34}{Red} boxes) are the baseline results. The 4rd row shows the bird-view results (\textcolor[RGB]{50,205,50}{Green} means the ground truth boxes). The IoU means the Intersection-over-Union between the predicted box and the corresponding ground-truth one and the uncertainty is the depth uncertainty $\sigma_d$ (best viewed in color.).}
\label{fig:uncertain}
\end{figure*}
\subsection{Ablation Study} 
To understand how much improvement each component provides, we perform ablation studies on the KITTI validation set for the Car category in Table~\ref{tab:ablation}. 


{\bf Effectiveness of the Base detector}.
In row (a), we establish a simple two-stage baseline, which detects objects at the RoI level. This model estimates depth directly by a depth head which has the same structure as other 3D detection heads, consisting of one 3x3 Conv layer with 256 channels, one averaged pooling layer and one fully-connected layer. To make the comparison fair, we let the depth head predict both value (mean) and uncertainty (standard deviation) and also utilize the Laplacian uncertainty loss function to train the model. 

To evaluate the effectiveness of our RoI-FM, we combine it for each RoI feature. The experiment (a$\rightarrow$b) clearly shows gains from this design, which proves location/size cues from the coordinate map and category information are crucial to monocular 3D detection.
Then, we introduce the GeP into depth estimation, which computes depth by estimated 2D/3D heights. Note that, here, the depth uncertainty is still estimated by a neural network head directly. The performances are shown in Table~\ref{tab:ablation} row (c). It can be seen that adding the GeP part improves performances in the experiment (b$\rightarrow$c)

\noindent
{\bf Comparison of Geometry Uncertainty Projection (GUP)}.
Some ablation study experiments for the GUPNet++ are related to the GUPNet~\cite{lu2021geometry}. So, to illustrate the ablation more clearly, we put the original ablation study of the GUPNet in lines (d)$\sim$(h) of Table~\ref{tab:ablation} here. To illustrate these ablations clearly, we provide a brief introduction of concerned parts here. Details could be found in the appendix.

\begin{figure*}[!ht]
\begin{center}
\includegraphics[width=1.0\linewidth]{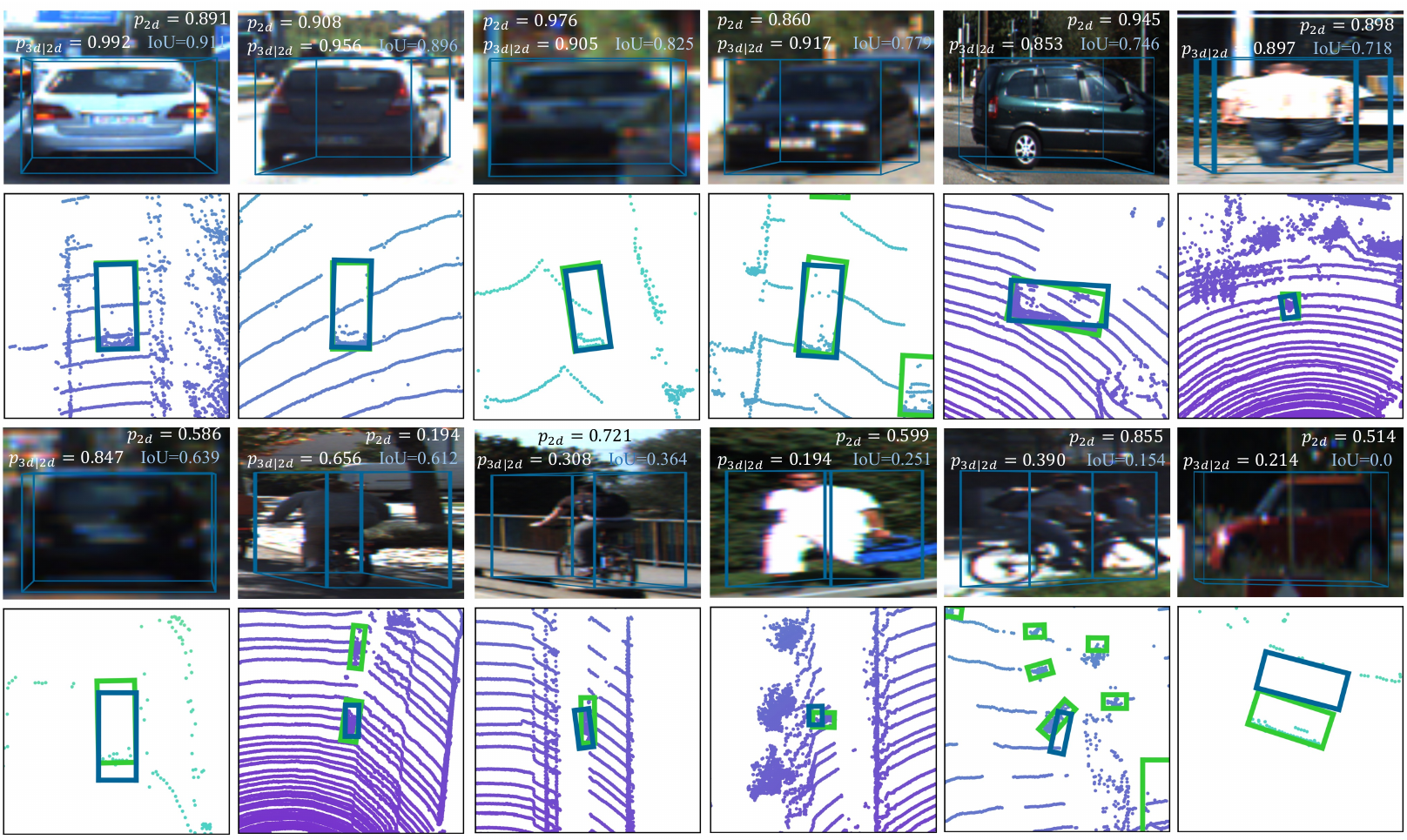}
\end{center}
\caption{The visualized examples of different scores, where $p_{2d}$ is the 2D detection score and $p_{3d|2d}$ is the conditioned 3D detection score. 1st and 3rd rows are the visualized bounding box while 2nd and 4th rows are corresponding bird-view results. \textcolor[RGB]{0,102,153}{Blue} boxes represents the box prediction and \textcolor[RGB]{50,205,50}{Green} boxes are the ground-truth boxes. }
\captionsetup{justification=centering}
\label{fig:vis_score}
\end{figure*}

The GUPNet ablations investigate the contribution for the Geometry-guided Uncertainty (GeU), vanilla Uncertainty-Confidence (UnC) and Hierarchical Task Learning (HTL). GeU derives the projected depth uncertainty as the following:
\begin{equation}
\label{eq:geu}
    \sigma_p = \frac{f\cdot \sigma_{3d}}{h_{2d}},
\end{equation}
where $h_{2d}$ is the 2D height estimation scalar. And UnC means computing 3d detection score as follows:
\begin{equation}
\label{eq:v_unc}
    p_{3d|2d} = exp(-\sigma_d).
\end{equation}
HTL is a scheme to combine loss term together as follows:
\begin{equation}
\begin{aligned}
     \mathcal{L}_{total} = \sum_{i\in \mathcal{T}} w_{i}(t)\cdot \mathcal{L}_{i},
\end{aligned}
\end{equation}
where $w_{i}(t)$, a time-dependent weighting term, controls the contribution of $i$-th loss term dynamically. 
The effectiveness of these three parts are evaluated, respectively. 

We evaluate the effectiveness of the UnC by combining this scheme with our base detector under different settings ((b) without GeP and (c) with GeP). By comparing settings (b$\rightarrow$d and c$\rightarrow$e), we can find the UnC part effectively and stably improves the overall performance, \eg 1.09\% improvement for (b$\rightarrow$d) and 0.78\% improvement for (c$\rightarrow$e) on 3D detection task under moderate level, demonstrating that uncertainty is a good indicator to reflect the 3D detection result and can provide a better score to improve the 3D detection reliability.
However, there is an interesting phenomenon that when we fix the UnC scheme, adding the GeP part leads to an accuracy drop in the experiment (d$\rightarrow$e, both of these experiments directly learn depth uncertainty as $\sigma_d$).
This proves our motivation that it is hard for the projection-based model to directly learn accurate uncertainty or confidence because of the error amplification. 

Now, we introduce the GeU strategy that utilizing Equation~\ref{eq:geu} to derive the depth uncertainty in row (f). Compared with (e), (f) shows that the geometry-guided uncertainty provides 2.13\% gains on the Moderate performance. And different from d$\rightarrow$e leading to performance drops, c$\rightarrow$f shows positive results that with the GeU, UnC also brings improvements to the GeP methods, reflecting that the GUP module can solve the difficulty of confidence learning in the projection-based model. 
The experimental results clearly demonstrate the effectiveness of the geometry modeling method for all metrics.

We also quantify the contribution of the proposed Hierarchical Task Learning (HTL) strategy by two groups of control experiments (e$\rightarrow$g and f$\rightarrow$h), and both of them confirm the efficacy of the proposed HTL (improving performances for all metrics, and about 2\% improvements for easy level), proving its ability to stabilize the training process. 

\begin{figure*}[!ht]
\begin{center}
\includegraphics[width=0.95\linewidth]{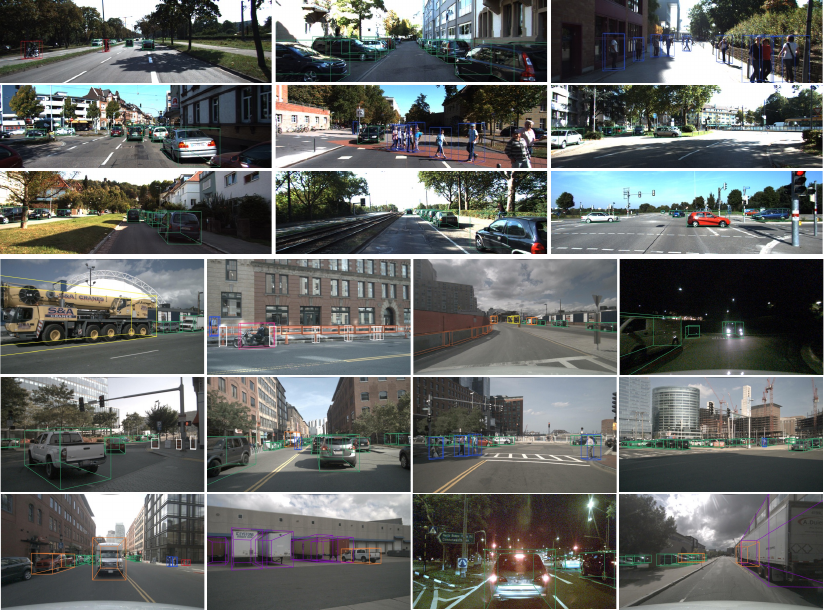}
\end{center}
\caption{Visualization of projected 3D Detection results of our GUPNet++ (best viewer in color.). The 1st$\sim$3th rows are KITTI results and 4th$\sim$6th are nuScenes one. For the shared categories across two datasets, car, pedestrian and cyclist/bicycle, we use \textcolor[RGB]{48,158,104}{Green}, \textcolor[RGB]{32,76,227}{Blue} and \textcolor[RGB]{216,30,44}{Red} to indicate them, respectively. And for nuScenes specific classes, we utilize \textcolor[RGB]{246,128,50}{Yellow Orange},\textcolor[RGB]{138,0,202}{Purple},\textcolor[RGB]{200,200,200}{White},\textcolor[RGB]{36,0,10}{Black},\textcolor[RGB]{233,224,50}{Yellow},\textcolor[RGB]{245,51,138}{Pink} and \textcolor[RGB]{142,78,52}{Red Orange} to represent truck, trailer, traffic cone, bus, construction vehicle, motorcycle and barrier, respectively.}
\label{fig:vis_res}
\end{figure*}
\subsubsection{GUPNet++}
\label{sec:ab_gupnet_plus}
The previous section shows the effectiveness of each part in GUPNet. Currently, we focus on evaluating the effectiveness of GUPNet++. In this section, we conduct experiments to explore the influence of the GUP++ module and the additional model augmentation schemes, respectively. 

\noindent
{\bf Comparison of Geometry Uncertainty Propagation (GUP++)}.
We conclude that our GUP++ module benefits the model by the following parts: new Geometry-guided Uncertainty (GeU++), IoU-guided Uncertainty-Confidence (IoUnC, Equation~\ref{eq:iaus}) and $\beta$-NLL loss function. 

We first evaluate the GeU++ which derives uncertainty by the uncertainty propagation shown in Equation~\ref{eq:geu++}. In row (i), we replace the original GeU of GUPNet (f) with the new GeU++. This model achieves 16.71\% AP on the car category of Moderate performance, showing 0.25\% gains compared with the original GUPNet and demonstrating the superiority of the new GeU++ scheme. After that, we delete the HTL scheme in (g) and it can be seen that the model achieves comparable results with (f), which is different from performance drops of removing the HTL in GUPNet. It proves that with the uncertainty propagation relationship in the GeU++, the model naturally adapts training of projection, leading to stabilized training without the need for any heuristic scheme. 

Further, we investigate the effectiveness of our new IoUnC scheme. The IoUnC is a parameter-free scheme, which is only utilized in the inference stage. We replace the vanilla UnC with the IoUnC and provide the results in row (k). It shows that the IoUnC improves the performance by about 1.2\%, 0.9\% and 0.54\% for Easy, Moderate and Hard metrics, respectively, proving the effectiveness of such a plug-and-play module. 

And finally, we explore the effectiveness of our new proposed $\beta$-NLL Laplacian loss function. Row (l) reflects the results that replacing the original NLL loss as the new $\beta$-NLL Laplacian loss, showing different level gains under different metrics and demonstrating the efficiency. 

\noindent
{\bf Model enhancements}. Here we provide the ablation about the techniques that we used for the model enhancements, including 3D Non-maximum Suppression (NMS3D) and the deformable convolution layer (DCN) in the DLA backbone.

From row (m) in Table~\ref{tab:ablation}, the NMS3D brings stable 1\%+ gains both on Moderate and Hard metrics, showing the effectiveness of such a scheme. Intuitively, compared with NMS on the 2D level (written as NMS2D), NMS3D is more flexible for 3D object detection. NMS2D only keeps local 2D detection results with the highest confidence. However, the feature that produces the highest confident 2D detection results may not provide the best 3D regression results, leading to the gap between the requirements of 3D object detection. The NMS3D could solve this gap well because the NMS operation is added to the 3D score. 

Further, we test the DCN operation in the DLA backbone and provide the results in the last row of Table~\ref{tab:ablation} (corresponding to the complete GUPNet++). The DCN operation provides other significant gains on all metrics. The benefit of the DCN can also be found in several recent monocular 3D object works~\cite{wang2021fcos3d,wang2022probabilistic}.

\subsection{Qualitative Results}
For further investigating the effectiveness of our GUPNet++. 
We show some cases and corresponding uncertainties of our GUPNet++, GUPNet and our baseline method. The results are shown in Figure~\ref{fig:uncertain}. We can see that our GUPNet++ predicts with high uncertainties for different bad cases including occlusion and far distance. And with the improvement of the prediction results, the uncertainty prediction of our method basically decreases. The original GUPNet also has similar trends but sometimes worse than the GUPNet++. And the baseline model gives unreliable uncertainty values in such cases, which shows the efficiency of our models.  

Except that, we provide additional visualization results of different scores of our GUPNet++, $p_{2d}$ and $p_{3d|2d}$ in Figure~\ref{fig:vis_score}. It can be seen that our $p_{3d|2d}$ could represent the quality of the 3D box estimation, which can provide a different view and compensate for the 2D score $p_{2d}$, leading to a more reliable final detection score.

Also, we show some detection results of our method. We first project the 3D bounding boxes generated by our method on the image plane to draw them on the scene image. The results are shown in Figure~\ref{fig:vis_res}.

\section{Conclusion}
In this paper, we propose GUPNet++ model, an updated version of GUPNet, to tackle the error amplification in the geometry projection process for monocular 3D object detection. It combines mathematical projection priors and the deep regression power together to compute more reliable uncertainty for each object, which not only be helpful for uncertainty-based learning but also can be used to compute accurate confidence in the testing stage by another novel IoU-guided Uncertainty-Confidence scheme. Extensive experiments validate the superior performance of the proposed algorithm in KITTI and nuScenes benchmarks, as well as the effectiveness of each component of the model.

\section{Acknowledgement}
This work was supported by the National Key R\&D Program of China, the Australian Medical Research Future Fund MRFAI000085, CRC-P Smart Material Recovery Facility (SMRF) – Curby Soft Plastics, and CRC-P ARIA - Bionic Visual-Spatial Prosthesis for the Blind*.


%

\ifCLASSOPTIONcaptionsoff
  \newpage
\fi



%
{\small
\bibliographystyle{unsrt}
\bibliography{bib}

\begin{thebibliography}{100}

\bibitem{meyer2019lasernet}
Gregory~P Meyer, Ankit Laddha, Eric Kee, Carlos Vallespi-Gonzalez, and Carl~K
  Wellington.
\newblock Lasernet: An efficient probabilistic 3d object detector for
  autonomous driving.
\newblock In {\em Proceedings of the IEEE/CVF Conference on Computer Vision and
  Pattern Recognition}, pages 12677--12686, 2019.

\bibitem{qi2019deep}
Charles~R Qi, Or~Litany, Kaiming He, and Leonidas~J Guibas.
\newblock Deep hough voting for 3d object detection in point clouds.
\newblock In {\em Proceedings of the IEEE/CVF International Conference on
  Computer Vision}, pages 9277--9286, 2019.

\bibitem{qin2019triangulation}
Zengyi Qin, Jinglu Wang, and Yan Lu.
\newblock Triangulation learning network: from monocular to stereo 3d object
  detection.
\newblock In {\em Proceedings of the IEEE/CVF Conference on Computer Vision and
  Pattern Recognition}, pages 7615--7623, 2019.

\bibitem{shi2019pointrcnn}
Shaoshuai Shi, Xiaogang Wang, and Hongsheng Li.
\newblock Pointrcnn: 3d object proposal generation and detection from point
  cloud.
\newblock In {\em Proceedings of the IEEE/CVF Conference on Computer Vision and
  Pattern Recognition}, pages 770--779, 2019.

\bibitem{shi2019part}
Shaoshuai Shi, Zhe Wang, Xiaogang Wang, and Hongsheng Li.
\newblock Part-a\({}^{\mbox{2}}\) net: 3d part-aware and aggregation neural
  network for object detection from point cloud.
\newblock {\em CoRR}, abs/1907.03670, 2019.

\bibitem{yang2019std}
Zetong Yang, Yanan Sun, Shu Liu, Xiaoyong Shen, and Jiaya Jia.
\newblock Std: Sparse-to-dense 3d object detector for point cloud.
\newblock In {\em Proceedings of the IEEE/CVF International Conference on
  Computer Vision}, pages 1951--1960, 2019.

\bibitem{yuan2021temporal}
Zhenxun Yuan, Xiao Song, Lei Bai, Zhe Wang, and Wanli Ouyang.
\newblock Temporal-channel transformer for 3d lidar-based video object
  detection for autonomous driving.
\newblock {\em IEEE Transactions on Circuits and Systems for Video Technology},
  32(4):2068--2078, 2021.

\bibitem{qin2019monogrnet}
Zengyi Qin, Jinglu Wang, and Yan Lu.
\newblock Monogrnet: A geometric reasoning network for monocular 3d object
  localization.
\newblock In {\em Proceedings of the AAAI Conference on Artificial
  Intelligence}, volume~33, pages 8851--8858, 2019.

\bibitem{weng2019monocular}
Xinshuo Weng and Kris Kitani.
\newblock Monocular 3d object detection with pseudo-lidar point cloud.
\newblock In {\em Proceedings of the IEEE/CVF International Conference on
  Computer Vision Workshops}, pages 0--0, 2019.

\bibitem{bao2020object}
Wentao Bao, Qi~Yu, and Yu~Kong.
\newblock Object-aware centroid voting for monocular 3d object detection.
\newblock {\em arXiv preprint arXiv:2007.09836}, 2020.

\bibitem{barabanau2019monocular}
I~Barabanau, A~Artemov, E~Burnaev, and V~Murashkin.
\newblock Monocular 3d object detection via geometric reasoning on keypoints.
\newblock In {\em VISIGRAPP 2020-Proceedings of the 15th International Joint
  Conference on Computer Vision, Imaging and Computer Graphics Theory and
  Applications}, pages 652--659, 2020.

\bibitem{cai2020monocular}
Yingjie Cai, Buyu Li, Zeyu Jiao, Hongsheng Li, Xingyu Zeng, and Xiaogang Wang.
\newblock Monocular 3d object detection with decoupled structured polygon
  estimation and height-guided depth estimation.
\newblock In {\em Proceedings of the AAAI Conference on Artificial
  Intelligence}, volume~34, pages 10478--10485, 2020.

\bibitem{ku2019monocular}
Jason Ku, Alex~D Pon, and Steven~L Waslander.
\newblock Monocular 3d object detection leveraging accurate proposals and shape
  reconstruction.
\newblock In {\em Proceedings of the IEEE/CVF Conference on Computer Vision and
  Pattern Recognition}, pages 11867--11876, 2019.

\bibitem{lu2021geometry}
Yan Lu, Xinzhu Ma, Lei Yang, Tianzhu Zhang, Yating Liu, Qi~Chu, Junjie Yan, and
  Wanli Ouyang.
\newblock Geometry uncertainty projection network for monocular 3d object
  detection.
\newblock In {\em Proceedings of the IEEE/CVF International Conference on
  Computer Vision}, pages 3111--3121, 2021.

\bibitem{blundell2015weight}
Charles Blundell, Julien Cornebise, Koray Kavukcuoglu, and Daan Wierstra.
\newblock Weight uncertainty in neural network.
\newblock In {\em International Conference on Machine Learning}, pages
  1613--1622. PMLR, 2015.

\bibitem{kendall2017uncertainties}
Alex Kendall and Yarin Gal.
\newblock What uncertainties do we need in bayesian deep learning for computer
  vision?
\newblock {\em arXiv preprint arXiv:1703.04977}, 2017.

\bibitem{he2018softer}
Yihui He, Xiangyu Zhang, Marios Savvides, and Kris Kitani.
\newblock Softer-nms: Rethinking bounding box regression for accurate object
  detection.
\newblock {\em CoRR}, abs/1809.08545, 2018.

\bibitem{gundavarapu2019structured}
Nitesh~B. Gundavarapu, Divyansh Srivastava, Rahul Mitra, Abhishek Sharma, and
  Arjun Jain.
\newblock Structured aleatoric uncertainty in human pose estimation.
\newblock In {\em {IEEE} Conference on Computer Vision and Pattern Recognition
  Workshops, {CVPR} Workshops 2019, Long Beach, CA, USA, June 16-20, 2019},
  pages 50--53. Computer Vision Foundation / {IEEE}, 2019.

\bibitem{dorta2018structured}
Garoe Dorta, Sara Vicente, Lourdes Agapito, Neill~DF Campbell, and Ivor
  Simpson.
\newblock Structured uncertainty prediction networks.
\newblock In {\em Proceedings of the IEEE conference on computer vision and
  pattern recognition}, pages 5477--5485, 2018.

\bibitem{li2021human}
Jiefeng Li, Siyuan Bian, Ailing Zeng, Can Wang, Bo~Pang, Wentao Liu, and Cewu
  Lu.
\newblock Human pose regression with residual log-likelihood estimation.
\newblock In {\em Proceedings of the IEEE/CVF international conference on
  computer vision}, pages 11025--11034, 2021.

\bibitem{rezende2015variational}
Danilo Rezende and Shakir Mohamed.
\newblock Variational inference with normalizing flows.
\newblock In {\em International conference on machine learning}, pages
  1530--1538. PMLR, 2015.

\bibitem{yu2019robust}
Tianyuan Yu, Da~Li, Yongxin Yang, Timothy~M Hospedales, and Tao Xiang.
\newblock Robust person re-identification by modelling feature uncertainty.
\newblock In {\em Proceedings of the IEEE/CVF international conference on
  computer vision}, pages 552--561, 2019.

\bibitem{chun2021probabilistic}
Sanghyuk Chun, Seong~Joon Oh, Rafael~Sampaio De~Rezende, Yannis Kalantidis, and
  Diane Larlus.
\newblock Probabilistic embeddings for cross-modal retrieval.
\newblock In {\em Proceedings of the IEEE/CVF Conference on Computer Vision and
  Pattern Recognition}, pages 8415--8424, 2021.

\bibitem{jin2020uncertainty}
Xin Jin, Cuiling Lan, Wenjun Zeng, and Zhibo Chen.
\newblock Uncertainty-aware multi-shot knowledge distillation for image-based
  object re-identification.
\newblock In {\em Proceedings of the AAAI Conference on Artificial
  Intelligence}, volume~34, pages 11165--11172, 2020.

\bibitem{kendall2018multi}
Alex Kendall, Yarin Gal, and Roberto Cipolla.
\newblock Multi-task learning using uncertainty to weigh losses for scene
  geometry and semantics.
\newblock In {\em Proceedings of the IEEE conference on computer vision and
  pattern recognition}, pages 7482--7491, 2018.

\bibitem{liu2019neural}
Chao Liu, Jinwei Gu, Kihwan Kim, Srinivasa~G Narasimhan, and Jan Kautz.
\newblock Neural rgb (r) d sensing: Depth and uncertainty from a video camera.
\newblock In {\em Proceedings of the IEEE/CVF Conference on Computer Vision and
  Pattern Recognition}, pages 10986--10995, 2019.

\bibitem{ding2020learning}
Mingyu Ding, Yuqi Huo, Hongwei Yi, Zhe Wang, Jianping Shi, Zhiwu Lu, and Ping
  Luo.
\newblock Learning depth-guided convolutions for monocular 3d object detection.
\newblock In {\em Proceedings of the IEEE/CVF Conference on Computer Vision and
  Pattern Recognition Workshops}, pages 1000--1001, 2020.

\bibitem{he2019mono3d++}
Tong He and Stefano Soatto.
\newblock Mono3d++: Monocular 3d vehicle detection with two-scale 3d hypotheses
  and task priors.
\newblock In {\em Proceedings of the AAAI Conference on Artificial
  Intelligence}, volume~33, pages 8409--8416, 2019.

\bibitem{kundu20183d}
Abhijit Kundu, Yin Li, and James~M Rehg.
\newblock 3d-rcnn: Instance-level 3d object reconstruction via
  render-and-compare.
\newblock In {\em Proceedings of the IEEE conference on computer vision and
  pattern recognition}, pages 3559--3568, 2018.

\bibitem{liu2019deep}
Lijie Liu, Jiwen Lu, Chunjing Xu, Qi~Tian, and Jie Zhou.
\newblock Deep fitting degree scoring network for monocular 3d object
  detection.
\newblock In {\em Proceedings of the IEEE/CVF Conference on Computer Vision and
  Pattern Recognition}, pages 1057--1066, 2019.

\bibitem{manhardt2019roi}
Fabian Manhardt, Wadim Kehl, and Adrien Gaidon.
\newblock Roi-10d: Monocular lifting of 2d detection to 6d pose and metric
  shape.
\newblock In {\em Proceedings of the IEEE/CVF Conference on Computer Vision and
  Pattern Recognition}, pages 2069--2078, 2019.

\bibitem{simonelli2019disentangling}
Andrea Simonelli, Samuel~Rota Bulo, Lorenzo Porzi, Manuel L{\'o}pez-Antequera,
  and Peter Kontschieder.
\newblock Disentangling monocular 3d object detection.
\newblock In {\em Proceedings of the IEEE/CVF International Conference on
  Computer Vision}, pages 1991--1999, 2019.

\bibitem{mousavian20173d}
Arsalan Mousavian, Dragomir Anguelov, John Flynn, and Jana Kosecka.
\newblock 3d bounding box estimation using deep learning and geometry.
\newblock In {\em Proceedings of the IEEE Conference on Computer Vision and
  Pattern Recognition}, pages 7074--7082, 2017.

\bibitem{chabot2017deep}
Florian Chabot, Mohamed Chaouch, Jaonary Rabarisoa, C{\'e}line Teuliere, and
  Thierry Chateau.
\newblock Deep manta: A coarse-to-fine many-task network for joint 2d and 3d
  vehicle analysis from monocular image.
\newblock In {\em Proceedings of the IEEE conference on computer vision and
  pattern recognition}, pages 2040--2049, 2017.

\bibitem{lian2022exploring}
Qing Lian, Botao Ye, Ruijia Xu, Weilong Yao, and Tong Zhang.
\newblock Exploring geometric consistency for monocular 3d object detection.
\newblock In {\em Proceedings of the IEEE/CVF Conference on Computer Vision and
  Pattern Recognition}, pages 1685--1694, 2022.

\bibitem{li2019gs3d}
Buyu Li, Wanli Ouyang, Lu~Sheng, Xingyu Zeng, and Xiaogang Wang.
\newblock Gs3d: An efficient 3d object detection framework for autonomous
  driving.
\newblock In {\em Proceedings of the IEEE/CVF Conference on Computer Vision and
  Pattern Recognition}, pages 1019--1028, 2019.

\bibitem{girshick2015fast}
Ross Girshick.
\newblock Fast r-cnn.
\newblock In {\em Proceedings of the IEEE international conference on computer
  vision}, pages 1440--1448, 2015.

\bibitem{brazil2019m3d}
Garrick Brazil and Xiaoming Liu.
\newblock M3d-rpn: Monocular 3d region proposal network for object detection.
\newblock In {\em Proceedings of the IEEE/CVF International Conference on
  Computer Vision}, pages 9287--9296, 2019.

\bibitem{kumar2022deviant}
Abhinav Kumar, Garrick Brazil, Enrique Corona, Armin Parchami, and Xiaoming
  Liu.
\newblock Deviant: Depth equivariant network for monocular 3d object detection.
\newblock In {\em European Conference on Computer Vision}, pages 664--683.
  Springer, 2022.

\bibitem{zhou2023monoatt}
Yunsong Zhou, Hongzi Zhu, Quan Liu, Shan Chang, and Minyi Guo.
\newblock Monoatt: Online monocular 3d object detection with adaptive token
  transformer.
\newblock In {\em Proceedings of the IEEE/CVF Conference on Computer Vision and
  Pattern Recognition}, pages 17493--17503, 2023.

\bibitem{ma2019accurate}
Xinzhu Ma, Zhihui Wang, Haojie Li, Pengbo Zhang, Wanli Ouyang, and Xin Fan.
\newblock Accurate monocular 3d object detection via color-embedded 3d
  reconstruction for autonomous driving.
\newblock In {\em Proceedings of the IEEE/CVF International Conference on
  Computer Vision}, pages 6851--6860, 2019.

\bibitem{qi2018frustum}
Charles~R Qi, Wei Liu, Chenxia Wu, Hao Su, and Leonidas~J Guibas.
\newblock Frustum pointnets for 3d object detection from rgb-d data.
\newblock In {\em Proceedings of the IEEE conference on computer vision and
  pattern recognition}, pages 918--927, 2018.

\bibitem{wang2019pseudo}
Yan Wang, Wei-Lun Chao, Divyansh Garg, Bharath Hariharan, Mark Campbell, and
  Kilian~Q Weinberger.
\newblock Pseudo-lidar from visual depth estimation: Bridging the gap in 3d
  object detection for autonomous driving.
\newblock In {\em Proceedings of the IEEE/CVF Conference on Computer Vision and
  Pattern Recognition}, pages 8445--8453, 2019.

\bibitem{ma2020rethinking}
Xinzhu Ma, Shinan Liu, Zhiyi Xia, Hongwen Zhang, Xingyu Zeng, and Wanli Ouyang.
\newblock Rethinking pseudo-lidar representation.
\newblock In {\em European Conference on Computer Vision}, pages 311--327.
  Springer, 2020.

\bibitem{simonelli2021we}
Andrea Simonelli, Samuel~Rota Bulo, Lorenzo Porzi, Peter Kontschieder, and
  Elisa Ricci.
\newblock Are we missing confidence in pseudo-lidar methods for monocular 3d
  object detection?
\newblock In {\em Proceedings of the IEEE/CVF International Conference on
  Computer Vision}, pages 3225--3233, 2021.

\bibitem{chen2022pseudo}
Yi-Nan Chen, Hang Dai, and Yong Ding.
\newblock Pseudo-stereo for monocular 3d object detection in autonomous
  driving.
\newblock In {\em Proceedings of the IEEE/CVF Conference on Computer Vision and
  Pattern Recognition}, pages 887--897, 2022.

\bibitem{reading2021categorical}
Cody Reading, Ali Harakeh, Julia Chae, and Steven~L Waslander.
\newblock Categorical depth distribution network for monocular 3d object
  detection.
\newblock {\em arXiv preprint arXiv:2103.01100}, 2021.

\bibitem{li2022diversity}
Zhuoling Li, Zhan Qu, Yang Zhou, Jianzhuang Liu, Haoqian Wang, and Lihui Jiang.
\newblock Diversity matters: Fully exploiting depth clues for reliable
  monocular 3d object detection.
\newblock In {\em Proceedings of the IEEE/CVF Conference on Computer Vision and
  Pattern Recognition}, pages 2791--2800, 2022.

\bibitem{peng2022did}
Liang Peng, Xiaopei Wu, Zheng Yang, Haifeng Liu, and Deng Cai.
\newblock Did-m3d: Decoupling instance depth for monocular 3d object detection.
\newblock In {\em European Conference on Computer Vision}, pages 71--88.
  Springer, 2022.

\bibitem{chen2020monopair}
Yongjian Chen, Lei Tai, Kai Sun, and Mingyang Li.
\newblock Monopair: Monocular 3d object detection using pairwise spatial
  relationships.
\newblock In {\em Proceedings of the IEEE/CVF Conference on Computer Vision and
  Pattern Recognition}, pages 12093--12102, 2020.

\bibitem{gu2022homography}
Jiaqi Gu, Bojian Wu, Lubin Fan, Jianqiang Huang, Shen Cao, Zhiyu Xiang, and
  Xian-Sheng Hua.
\newblock Homography loss for monocular 3d object detection.
\newblock In {\em Proceedings of the IEEE/CVF Conference on Computer Vision and
  Pattern Recognition}, pages 1080--1089, 2022.

\bibitem{wang2022probabilistic}
Tai Wang, ZHU Xinge, Jiangmiao Pang, and Dahua Lin.
\newblock Probabilistic and geometric depth: Detecting objects in perspective.
\newblock In {\em Conference on Robot Learning}, pages 1475--1485. PMLR, 2022.

\bibitem{wang2021fcos3d}
Tai Wang, Xinge Zhu, Jiangmiao Pang, and Dahua Lin.
\newblock Fcos3d: Fully convolutional one-stage monocular 3d object detection.
\newblock In {\em Proceedings of the IEEE/CVF International Conference on
  Computer Vision}, pages 913--922, 2021.

\bibitem{zhang2022monodetr}
Renrui Zhang, Han Qiu, Tai Wang, Ziyu Guo, Xuanzhuo Xu, Yu~Qiao, Peng Gao, and
  Hongsheng Li.
\newblock Monodetr: depth-guided transformer for monocular 3d object detection.
\newblock {\em arXiv preprint arXiv:2203.13310}, 2022.

\bibitem{carion2020end}
Nicolas Carion, Francisco Massa, Gabriel Synnaeve, Nicolas Usunier, Alexander
  Kirillov, and Sergey Zagoruyko.
\newblock End-to-end object detection with transformers.
\newblock In {\em European conference on computer vision}, pages 213--229.
  Springer, 2020.

\bibitem{zhou2021monocular}
Yunsong Zhou, Yuan He, Hongzi Zhu, Cheng Wang, Hongyang Li, and Qinhong Jiang.
\newblock Monocular 3d object detection: An extrinsic parameter free approach.
\newblock In {\em Proceedings of the IEEE/CVF Conference on Computer Vision and
  Pattern Recognition}, pages 7556--7566, 2021.

\bibitem{zhou2021monoef}
Yunsong Zhou, Yuan He, Hongzi Zhu, Cheng Wang, Hongyang Li, and Qinhong Jiang.
\newblock Monoef: Extrinsic parameter free monocular 3d object detection.
\newblock {\em IEEE Transactions on Pattern Analysis and Machine Intelligence},
  44(12):10114--10128, 2021.

\bibitem{zhou2022mogde}
Yunsong Zhou, Quan Liu, Hongzi Zhu, Yunzhe Li, Shan Chang, and Minyi Guo.
\newblock Mogde: Boosting mobile monocular 3d object detection with ground
  depth estimation.
\newblock {\em Advances in Neural Information Processing Systems},
  35:2033--2045, 2022.

\bibitem{chen2021monorun}
Hansheng Chen, Yuyao Huang, Wei Tian, Zhong Gao, and Lu~Xiong.
\newblock Monorun: Monocular 3d object detection by reconstruction and
  uncertainty propagation.
\newblock In {\em Proceedings of the IEEE/CVF Conference on Computer Vision and
  Pattern Recognition}, pages 10379--10388, 2021.

\bibitem{shi2021geometry}
Xuepeng Shi, Qi~Ye, Xiaozhi Chen, Chuangrong Chen, Zhixiang Chen, and Tae-Kyun
  Kim.
\newblock Geometry-based distance decomposition for monocular 3d object
  detection.
\newblock In {\em Proceedings of the IEEE/CVF International Conference on
  Computer Vision}, pages 15172--15181, 2021.

\bibitem{shi2023multivariate}
Xuepeng Shi, Zhixiang Chen, and Tae-Kyun Kim.
\newblock Multivariate probabilistic monocular 3d object detection.
\newblock In {\em Proceedings of the IEEE/CVF Winter Conference on Applications
  of Computer Vision}, pages 4281--4290, 2023.

\bibitem{zhou2019objects}
Xingyi Zhou, Dequan Wang, and Philipp Kr{\"{a}}henb{\"{u}}hl.
\newblock Objects as points.
\newblock {\em CoRR}, abs/1904.07850, 2019.

\bibitem{he2017mask}
Kaiming He, Georgia Gkioxari, Piotr Doll{\'a}r, and Ross Girshick.
\newblock Mask r-cnn.
\newblock In {\em Proceedings of the IEEE international conference on computer
  vision}, pages 2961--2969, 2017.

\bibitem{dijk2019neural}
Tom~van Dijk and Guido~de Croon.
\newblock How do neural networks see depth in single images?
\newblock In {\em Proceedings of the IEEE/CVF International Conference on
  Computer Vision}, pages 2183--2191, 2019.

\bibitem{kitti}
Andreas Geiger, Philip Lenz, and Raquel Urtasun.
\newblock Are we ready for autonomous driving? the kitti vision benchmark
  suite.
\newblock In {\em 2012 IEEE Conference on Computer Vision and Pattern
  Recognition}, pages 3354--3361. IEEE, 2012.

\bibitem{caesar2020nuscenes}
Holger Caesar, Varun Bankiti, Alex~H Lang, Sourabh Vora, Venice~Erin Liong,
  Qiang Xu, Anush Krishnan, Yu~Pan, Giancarlo Baldan, and Oscar Beijbom.
\newblock nuscenes: A multimodal dataset for autonomous driving.
\newblock In {\em Proceedings of the IEEE/CVF conference on computer vision and
  pattern recognition}, pages 11621--11631, 2020.

\bibitem{tatarchenko2019single}
Maxim Tatarchenko, Stephan~R Richter, Ren{\'e} Ranftl, Zhuwen Li, Vladlen
  Koltun, and Thomas Brox.
\newblock What do single-view 3d reconstruction networks learn?
\newblock In {\em Proceedings of the IEEE/CVF conference on computer vision and
  pattern recognition}, pages 3405--3414, 2019.

\bibitem{lin2017focal}
Tsung-Yi Lin, Priya Goyal, Ross Girshick, Kaiming He, and Piotr Doll{\'a}r.
\newblock Focal loss for dense object detection.
\newblock In {\em Proceedings of the IEEE international conference on computer
  vision}, pages 2980--2988, 2017.

\bibitem{law2018cornernet}
Hei Law and Jia Deng.
\newblock Cornernet: Detecting objects as paired keypoints.
\newblock In {\em Proceedings of the European conference on computer vision
  (ECCV)}, pages 734--750, 2018.

\bibitem{wiki:Propagation_of_uncertainty}
Wikipedia.
\newblock {Propagation of uncertainty} --- {W}ikipedia{,} the free
  encyclopedia.
\newblock
  \url{http://en.wikipedia.org/w/index.php?title=Propagation\%20of\%20uncertainty&oldid=1167031967},
  2023.
\newblock [Online; accessed 31-July-2023].

\bibitem{seitzer2022pitfalls}
Maximilian Seitzer, Arash Tavakoli, Dimitrije Antic, and Georg Martius.
\newblock On the pitfalls of heteroscedastic uncertainty estimation with
  probabilistic neural networks.
\newblock In {\em The Tenth International Conference on Learning
  Representations, {ICLR} 2022, Virtual Event, April 25-29, 2022}.
  OpenReview.net, 2022.

\bibitem{fang2022deepvp}
Zhencheng Fang, Tao Feng, Hongwei Zhou, and Muxuan Chen.
\newblock Deepvp: Identification and classification of phage virion proteins
  using deep learning.
\newblock {\em GigaScience}, 11:giac076, 2022.

\bibitem{da3dnet}
Xiaoqing Ye, Liang Du, Yifeng Shi, Yingying Li, Xiao Tan, Jianfeng Feng, Errui
  Ding, and Shilei Wen.
\newblock Monocular 3d object detection via feature domain adaptation.
\newblock In {\em Computer Vision--ECCV 2020: 16th European Conference,
  Glasgow, UK, August 23--28, 2020, Proceedings, Part IX 16}, pages 17--34.
  Springer, 2020.

\bibitem{huang2022monodtr}
Kuan-Chih Huang, Tsung-Han Wu, Hung-Ting Su, and Winston~H Hsu.
\newblock Monodtr: Monocular 3d object detection with depth-aware transformer.
\newblock In {\em Proceedings of the IEEE/CVF Conference on Computer Vision and
  Pattern Recognition}, pages 4012--4021, 2022.

\bibitem{hong2022cross}
Yu~Hong, Hang Dai, and Yong Ding.
\newblock Cross-modality knowledge distillation network for monocular 3d object
  detection.
\newblock In {\em European Conference on Computer Vision}, pages 87--104.
  Springer, 2022.

\bibitem{wu2023monopgc}
Zizhang Wu, Yuanzhu Gan, Lei Wang, Guilian Chen, and Jian Pu.
\newblock Monopgc: Monocular 3d object detection with pixel geometry contexts.
\newblock {\em arXiv preprint arXiv:2302.10549}, 2023.

\bibitem{brazil2020kinematic}
Garrick Brazil, Gerard Pons-Moll, Xiaoming Liu, and Bernt Schiele.
\newblock Kinematic 3d object detection in monocular video.
\newblock In {\em European Conference on Computer Vision}, pages 135--152.
  Springer, 2020.

\bibitem{chong2022monodistill}
Zhiyu Chong, Xinzhu Ma, Hong Zhang, Yuxin Yue, Haojie Li, Zhihui Wang, and
  Wanli Ouyang.
\newblock Monodistill: Learning spatial features for monocular 3d object
  detection.
\newblock In {\em The Tenth International Conference on Learning
  Representations, {ICLR} 2022, Virtual Event, April 25-29, 2022}.
  OpenReview.net, 2022.

\bibitem{liu2021autoshape}
Zongdai Liu, Dingfu Zhou, Feixiang Lu, Jin Fang, and Liangjun Zhang.
\newblock Autoshape: Real-time shape-aware monocular 3d object detection.
\newblock In {\em Proceedings of the IEEE/CVF International Conference on
  Computer Vision}, pages 15641--15650, 2021.

\bibitem{ur3d}
Xuepeng Shi, Zhixiang Chen, and Tae-Kyun Kim.
\newblock Distance-normalized unified representation for monocular 3d object
  detection.
\newblock In {\em European Conference on Computer Vision}, pages 91--107.
  Springer, 2020.

\bibitem{smoke}
Zechen Liu, Zizhang Wu, and Roland T{\'o}th.
\newblock Smoke: single-stage monocular 3d object detection via keypoint
  estimation.
\newblock In {\em Proceedings of the IEEE/CVF Conference on Computer Vision and
  Pattern Recognition Workshops}, pages 996--997, 2020.

\bibitem{li2020rtm3d}
Peixuan Li, Huaici Zhao, Pengfei Liu, and Feidao Cao.
\newblock Rtm3d: Real-time monocular 3d detection from object keypoints for
  autonomous driving.
\newblock {\em arXiv preprint arXiv:2001.03343}, 2, 2020.

\bibitem{movi3d}
Andrea Simonelli, Samuel~Rota Bul{\`o}, Lorenzo Porzi, Elisa Ricci, and Peter
  Kontschieder.
\newblock Towards generalization across depth for monocular 3d object
  detection.
\newblock {\em arXiv preprint arXiv:1912.08035}, 2019.

\bibitem{rukhovich2022imvoxelnet}
Danila Rukhovich, Anna Vorontsova, and Anton Konushin.
\newblock Imvoxelnet: Image to voxels projection for monocular and multi-view
  general-purpose 3d object detection.
\newblock In {\em Proceedings of the IEEE/CVF Winter Conference on Applications
  of Computer Vision}, pages 2397--2406, 2022.

\bibitem{rarnet}
Lijie Liu, Chufan Wu, Jiwen Lu, Lingxi Xie, Jie Zhou, and Qi~Tian.
\newblock Reinforced axial refinement network for monocular 3d object
  detection.
\newblock In {\em European Conference on Computer Vision}, pages 540--556.
  Springer, 2020.

\bibitem{zhang2021objects}
Yunpeng Zhang, Jiwen Lu, and Jie Zhou.
\newblock Objects are different: Flexible monocular 3d object detection.
\newblock In {\em Proceedings of the IEEE/CVF Conference on Computer Vision and
  Pattern Recognition}, pages 3289--3298, 2021.

\bibitem{li2022densely}
Yingyan Li, Yuntao Chen, Jiawei He, and Zhaoxiang Zhang.
\newblock Densely constrained depth estimator for monocular 3d object
  detection.
\newblock In {\em European Conference on Computer Vision}, pages 718--734.
  Springer, 2022.

\bibitem{mono3d}
Xiaozhi Chen, Kaustav Kundu, Ziyu Zhang, Huimin Ma, Sanja Fidler, and Raquel
  Urtasun.
\newblock Monocular 3d object detection for autonomous driving.
\newblock In {\em Proceedings of the IEEE Conference on Computer Vision and
  Pattern Recognition (CVPR)}, June 2016.

\bibitem{mv3d}
Xiaozhi Chen, Huimin Ma, Ji~Wan, Bo~Li, and Tian Xia.
\newblock Multi-view 3d object detection network for autonomous driving.
\newblock In {\em Proceedings of the IEEE Conference on Computer Vision and
  Pattern Recognition (CVPR)}, July 2017.

\bibitem{zhou2020tracking}
Xingyi Zhou, Vladlen Koltun, and Philipp Kr{\"a}henb{\"u}hl.
\newblock Tracking objects as points.
\newblock In {\em European conference on computer vision}, pages 474--490.
  Springer, 2020.

\bibitem{yu2018deep}
Fisher Yu, Dequan Wang, Evan Shelhamer, and Trevor Darrell.
\newblock Deep layer aggregation.
\newblock In {\em Proceedings of the IEEE conference on computer vision and
  pattern recognition}, pages 2403--2412, 2018.

\bibitem{dai2017deformable}
Jifeng Dai, Haozhi Qi, Yuwen Xiong, Yi~Li, Guodong Zhang, Han Hu, and Yichen
  Wei.
\newblock Deformable convolutional networks.
\newblock In {\em Proceedings of the IEEE international conference on computer
  vision}, pages 764--773, 2017.

\bibitem{newell2016stacked}
Alejandro Newell, Kaiyu Yang, and Jia Deng.
\newblock Stacked hourglass networks for human pose estimation.
\newblock In {\em Computer Vision--ECCV 2016: 14th European Conference,
  Amsterdam, The Netherlands, October 11-14, 2016, Proceedings, Part VIII 14},
  pages 483--499. Springer, 2016.

\bibitem{ma2021delving}
Xinzhu Ma, Yinmin Zhang, Dan Xu, Dongzhan Zhou, Shuai Yi, Haojie Li, and Wanli
  Ouyang.
\newblock Delving into localization errors for monocular 3d object detection.
\newblock In {\em Proceedings of the IEEE/CVF Conference on Computer Vision and
  Pattern Recognition}, pages 4721--4730, 2021.

\bibitem{zhang2021learning}
Yinmin Zhang, Xinzhu Ma, Shuai Yi, Jun Hou, Zhihui Wang, Wanli Ouyang, and Dan
  Xu.
\newblock Learning geometry-guided depth via projective modeling for monocular
  3d object detection.
\newblock {\em CoRR}, abs/2107.13931, 2021.

\bibitem{chen2016monocular}
Xiaozhi Chen, Kaustav Kundu, Ziyu Zhang, Huimin Ma, Sanja Fidler, and Raquel
  Urtasun.
\newblock Monocular 3d object detection for autonomous driving.
\newblock In {\em Proceedings of the IEEE conference on computer vision and
  pattern recognition}, pages 2147--2156, 2016.

\bibitem{roddick2018orthographic}
Thomas Roddick, Alex Kendall, and Roberto Cipolla.
\newblock Orthographic feature transform for monocular 3d object detection.
\newblock In {\em 30th British Machine Vision Conference 2019, {BMVC} 2019,
  Cardiff, UK, September 9-12, 2019}, page 285. {BMVA} Press, 2019.

\bibitem{liu2022petr}
Yingfei Liu, Tiancai Wang, Xiangyu Zhang, and Jian Sun.
\newblock Petr: Position embedding transformation for multi-view 3d object
  detection.
\newblock In {\em European Conference on Computer Vision}, pages 531--548.
  Springer, 2022.

\bibitem{he2016deep}
Kaiming He, Xiangyu Zhang, Shaoqing Ren, and Jian Sun.
\newblock Deep residual learning for image recognition.
\newblock In {\em Proceedings of the IEEE conference on computer vision and
  pattern recognition}, pages 770--778, 2016.

\bibitem{huang2021bevdet}
Junjie Huang, Guan Huang, Zheng Zhu, and Dalong Du.
\newblock Bevdet: High-performance multi-camera 3d object detection in
  bird-eye-view.
\newblock {\em CoRR}, abs/2112.11790, 2021.

\bibitem{liu2021swin}
Ze~Liu, Yutong Lin, Yue Cao, Han Hu, Yixuan Wei, Zheng Zhang, Stephen Lin, and
  Baining Guo.
\newblock Swin transformer: Hierarchical vision transformer using shifted
  windows.
\newblock In {\em Proceedings of the IEEE/CVF international conference on
  computer vision}, pages 10012--10022, 2021.

\bibitem{zou2023diffbev}
Jiayu Zou, Kun Tian, Zheng Zhu, Yun Ye, and Xingang Wang.
\newblock Diffbev: Conditional diffusion model for bird’s eye view
  perception.
\newblock In {\em Proceedings of the AAAI Conference on Artificial
  Intelligence}, volume~38, pages 7846--7854, 2024.

\bibitem{zhao2016loss}
Hang Zhao, Orazio Gallo, Iuri Frosio, and Jan Kautz.
\newblock Loss functions for image restoration with neural networks.
\newblock {\em IEEE Transactions on computational imaging}, 3(1):47--57, 2016.

\bibitem{lim2017enhanced}
Bee Lim, Sanghyun Son, Heewon Kim, Seungjun Nah, and Kyoung Mu~Lee.
\newblock Enhanced deep residual networks for single image super-resolution.
\newblock In {\em Proceedings of the IEEE conference on computer vision and
  pattern recognition workshops}, pages 136--144, 2017.

\bibitem{rashid2017interspecies}
Maheen Rashid, Xiuye Gu, and Yong Jae~Lee.
\newblock Interspecies knowledge transfer for facial keypoint detection.
\newblock In {\em Proceedings of the IEEE Conference on Computer Vision and
  Pattern Recognition}, pages 6894--6903, 2017.

\bibitem{bishop2006pattern}
Christopher~M Bishop and Nasser~M Nasrabadi.
\newblock {\em Pattern recognition and machine learning}, volume~4.
\newblock Springer, 2006.

\bibitem{morerio2017curriculum}
Pietro Morerio, Jacopo Cavazza, Riccardo Volpi, Ren{\'e} Vidal, and Vittorio
  Murino.
\newblock Curriculum dropout.
\newblock In {\em Proceedings of the IEEE International Conference on Computer
  Vision}, pages 3544--3552, 2017.

\bibitem{chen2018gradnorm}
Zhao Chen, Vijay Badrinarayanan, Chen-Yu Lee, and Andrew Rabinovich.
\newblock Gradnorm: Gradient normalization for adaptive loss balancing in deep
  multitask networks.
\newblock In {\em International Conference on Machine Learning}, pages
  794--803. PMLR, 2018.

\bibitem{zhang2014facial}
Zhanpeng Zhang, Ping Luo, Chen~Change Loy, and Xiaoou Tang.
\newblock Facial landmark detection by deep multi-task learning.
\newblock In {\em European conference on computer vision}, pages 94--108.
  Springer, 2014.

\end{thebibliography}
}
%





\clearpage
\newpage
\appendices
\section{Probabilistic View of the IoU-guided Uncertainty Confidence}
In this appendix section, we provide the probabilistic insight of the proposed IoU-guided Uncertainty Confidence (IoUnC) scheme. 

We first provide a brief review of the probabilistic view of our proposed GUPNet++. The depth output of our GUPNet++ consists of a mean $\mu_d$ and an associated uncertainty $\sigma_d$. These parameters are organized as a depth distribution indicating {\bf probability that the target depth is located at each potential location}. During training, the uncertainty optimization, as shown in Equation 12 in the original paper, guides the model to align the highest probability for ground-truth depth location. During inference, we choose $\mu_d$ as the final depth value for 3D bounding box $B^p$ because $\mu_d$ has the highest probability. However, the model often does not provide 100\% certain for $\mu_d$ so the depth distribution said that the target depth still has non-zero probabilities to located at other locations. 

If the target depth is located at $x$, having a drift from the highest probability $\mu_d$, what will happen? Intuitively, in object detection, if $x$ is not quite far away from $\mu_d$, we still believe $\mu_d$ is a good prediction. 
From this insight, we define the confidence of $\mu_d$ as the probability that $\mu_d$ is seen as a good prediction. This goal can be achieved as follows, same to Equation 18 in the original paper:
\begin{equation}
\label{eq:cdf2}
\begin{aligned}
p_{3d|2d}=\int_{\mu_d-\Delta d}^{\mu_d+\Delta d} p(D=x|\mu_d,\sigma_d) dx.
\end{aligned}
\end{equation}
This equation can be understood as that we traverse all target depth candidates that are not far away from $\mu_d$ and accumulate their probabilities. The $\Delta d$ in this equation is the maximal bound of accepted depth drift. If target depth $x$ is not in the range of $[\mu_d-\Delta d,\mu_d+\Delta d]$, $\mu_d$ is not good. 

Now, we go further about this idea. We expand the probability of $\mu_d$ being a good predicted depth to the probability of $B^p$, with $\mu_d$ as its depth, being a good predicted box. In other words, we finally define confidence as the probability that $B^p$ is a true positive predicted box. To make Equation~\ref{eq:cdf2} meet this expand, the definition of $\Delta d$ is modified as the maximal accepted depth drift between the target box and the predicted box. If the drift is larger than $\Delta d$, $B^p$ will become a false positive prediction. Based on that, $\Delta d$ can be derived by the object detection criteria as follows:
\begin{equation}
\label{eq:delta_d_define}
\begin{aligned} \forall x \in [\mu_d-\Delta d,\mu_d+\Delta d],\ \text{IoU}(B^p(d=x),B^p)\geq th,
\end{aligned}
\end{equation}
where $B^p(d=x)$ means manually set the depth of $B^p$ as $x$. This equation means that if the target depth $x$ is located in the range of $[\mu_d-\Delta d,\mu_d+\Delta d]$, the IoU between the predicted box $B^p$ and the target box $B^p(d=x)$ will be larger than the threshold $th$. Note that we utilize IoU to determine whether a box is true positive or not. Finally, Equation~\ref{eq:delta_d_define} can be derived into Equation 17 in the original paper. 

In conclusion, the IoUnC can be seen as an estimation of {\bf the probability that $B^p$ is a true positive predicted box}. $th$ is set as 0.7, which is widely used in several detection benchmarks. Please note that the specific value of $th$ would not affect the detection performance because it can only affect the absolute scale of $p_{3d|2d}$ but cannot change the relative ranking relationships of different samples. 

\section{Experiments under different $th$ of IoUnC}
We test performance under different $th$ in IoUnC and the results are shown in Table~\ref{tab:app_th}. 
It can be seen that the best results are archived at $th=0.7$. So we set it as the final hyper-parameter value. 
\begin{table}[!ht]
\centering
\fontsize{8}{10}\selectfont
\caption{{\bf Performances under different $th$ in IoUnC on the Car category of the KITTI \emph{validation} set. }  }
\label{tab:app_th}
\begin{tabular}{c||cccccc}
\toprule
$th$ value&0.0&0.1&0.3&0.5&0.7&0.9\cr\hline
AP40 (Easy)&26.37&26.43&26.84&27.49&29.03&27.83\cr
AP40 (Moderate)&19.23&19.29&19.51&19.82&20.45&19.41\cr
AP40 (Hard)&16.73&16.87&17.09&17.48&17.89&17.11\cr
\bottomrule  
\end{tabular}
\end{table}

\section{Necessity of the depth bias ($\mu_b,\sigma_b$)}
\label{sec:app_depth_bias}
\begin{figure}[ht]
    \centering
    \includegraphics[width=0.8\linewidth]{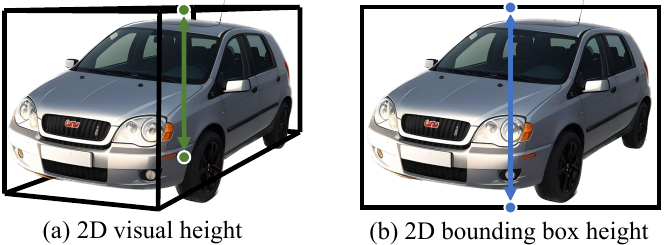}
    \caption{Visualization of two kind of $h_{2d}$.}
    \label{fig:enter-label}
\end{figure}

The necessity of the added bias is caused by the implementation of the projected depth. In the projection process, written as $d_p=f\cdot h_{3d}/h_{2d}$, $h_{2d}$ is implemented by 2D bounding box height instead of 2D visual height. The 2D bounding box height is larger than the 2D visual height as shown in Figure~\ref{fig:enter-label}.
Directly utilizing 2D bounding box height in projection would lead to a relatively smaller depth than the real depth value. So, $d_p$ with parameters ($\mu_p$, $\sigma_p$) alone cannot provide accurate depth. We add a biased term to modify this gap for both depth value and its uncertainty. 

\section{The proof of Equation 10}
The uncertainty propagation theory means: For a function $y=f(x_1, x_2, ..., x_n)$, if $x_1, x_2, ..., x_n$ are independent of each other, the uncertainty of $y$ can be derived as follows:
\begin{equation}
    \sigma_y = \sqrt{\left(\frac{\partial y}{\partial x_1}\right)^2\sigma_1^2 + \left(\frac{\partial y}{\partial x_2}\right)^2\sigma_2^2 + ... + \left(\frac{\partial y}{\partial x_n}\right)^2\sigma_n^2},
\end{equation}
where $\sigma_y$ is the uncertainty of $y$ and $\sigma^x_1, \sigma^x_2, ..., \sigma^x_n$ are the input uncertainties, respectively. So, for the depth derivation as shown in the following:
\begin{equation}
    d_p=\frac{f\cdot h_{3d}}{h_{2d}},
\end{equation}
the uncertainty for $d_p$, written as $\sigma_p$, can be computed as following:
\begin{equation}
\begin{aligned}
    \sigma_p &= \sqrt{\left(\frac{\partial d_p}{\partial h_{3d}}\right)^2\sigma_{3d}^2 + \left(\frac{\partial d_p}{\partial h_{2d}}\right)^2\sigma_{2d}^2}\\
    &=\sqrt{\left(\frac{f}{h_{2d}}\right)^2\sigma_{3d}^2 + \left(-\frac{f\cdot h_{3d}}{h_{2d}^2}\right)^2\sigma_{2d}^2}\\
\end{aligned}
\end{equation}
Substitute specific values, $h_{3d}=\mu_{3d}$ and $h_{2d}=\mu_{2d}$, then:
\begin{equation}
\begin{aligned}
    \sigma_p &=\sqrt{\left(\frac{f}{\mu_{2d}}\right)^2\sigma_{3d}^2 + \left(-\frac{f\cdot \mu_{3d}}{\mu_{2d}^2}\right)^2\sigma_{2d}^2}\\
    &=\sqrt{\frac{f^2}{\mu_{2d}^2}\sigma_{3d}^2 + \frac{f^2\cdot \mu_{3d}^2}{\mu_{2d}^4}\sigma_{2d}^2}\\
    &=\sqrt{\frac{f^2\cdot \mu_{3d}^2}{\mu_{2d}^2}\cdot(\frac{\sigma_{3d}^2}{\mu_{3d}^2} + \frac{\sigma_{2d}^2}{\mu_{2d}^2})}\\
    &=\mu_p\cdot\sqrt{\frac{\sigma_{3d}^2}{\mu_{3d}^2} + \frac{\sigma_{2d}^2}{\mu_{2d}^2}},\ \text{where}\ \mu_p = \frac{f\cdot \mu_{3d}}{\mu_{2d}}.
\end{aligned}
\end{equation}
Proof completed.

\section{The error of 2D and 3D height estimations}
\begin{table}[!ht]
\centering
\fontsize{8}{10}\selectfont
\caption{{\bf Height estimation absolute error and 2D detection mAP on the KITTI \emph{validation} set.}  }
\label{tab:sup_height_error}
\begin{tabular}{c||ccc|c}
\toprule
&Car&Pedestrian&Cyclist&Overall\cr\hline
$h_{2d}$ absolute error (pixel)&3.118&7.951&6.806&3.959\cr
$h_{3d}$ absolute error (meter)&0.083&0.084&0.089&0.083\cr\hline
2D detection mAP (Easy)&93.38&39.66&45.53&59.52\cr
2D detection mAP (Moderate)&83.13&32.80&26.53&47.49\cr
2D detection mAP (Hard)&76.65&28.38&25.09&43.37\cr
\bottomrule  
\end{tabular}
\end{table}
We evaluate the 2D height and 3D height estimation accuracy on the KITTI validation set. We report them with the 2D detection in Table~\ref{tab:sup_height_error}. The mean 2D height error is 3.959 pixels and the 3D height error is 0.083 meters. These results support our motivation. For the 3D height, its error is small but its influence on the depth is critical. For the 2D height, although the estimation error is satisfactory to the 2D detection, its magnitude is still non-neglectable for the depth projection. 

\section{Analysis of Distribution Choice for Regressions}
For the main regression topics, 2D height, 3D height and depth, in the GUPNet++, we have assumed them following Laplacian distributions. Here, we try to provide some analysis and insights for such choices. 

The Laplacian distribution assumption is aimed at training regression under the L1 loss function. If we utilize the Gaussian distribution assumption, the training loss will become L2. Their formulations are as follows:
\begin{equation}
    \begin{aligned}
        P^{\text{Lap}}(Y)=\frac{1}{\sqrt2\sigma}\exp\left(-\frac{\sqrt2|y-\mu|}{\sigma}\right)&\rightarrow\\\mathcal{L}^{\text{Lap}}_{\text{NLL}} = \frac{\sqrt2}{\sigma}\underbrace{|y-\mu|}_{\text{L1 term}}+\log\sigma,\\
        P^{\text{Gau}}(Y)=\frac{1}{\sqrt{2\pi}\sigma}\exp\left(-\frac{|y-\mu|^2}{2\sigma^2}\right)&\rightarrow\\\mathcal{L}^{\text{Gau}}_{\text{NLL}} = \frac{1}{2\sigma^2}\underbrace{(y-\mu)^2}_{\text{L2 term}}+\log\sigma,
    \end{aligned}
\end{equation}
We try the Gaussian setting on the GUPNet++, and its AP40 performance on the Car category on the KITTI validation set drops to 24.13\%, 16.34\%, and 14.22\% for the easy, moderate, and hard settings, respectively. We also try the Gaussian setting on the GUPNet and our baseline model. It stably leads to performance drops. We conclude the reason for this phenomenon as follows:

\noindent
1). L1 loss has been proved that can achieve better results in accurate regression tasks in computer vision~\cite{zhao2016loss,lim2017enhanced,rashid2017interspecies}. The main reason is that the L1 loss has larger loss values than the L2 under small range error, which makes the model fit better than the L2 loss. The monocular 3D object detection actually requires accurate regression, so we choose the L1-related loss rather than the L2-related one.  

\begin{figure*}
    \centering
    \includegraphics[width=1.0\linewidth]{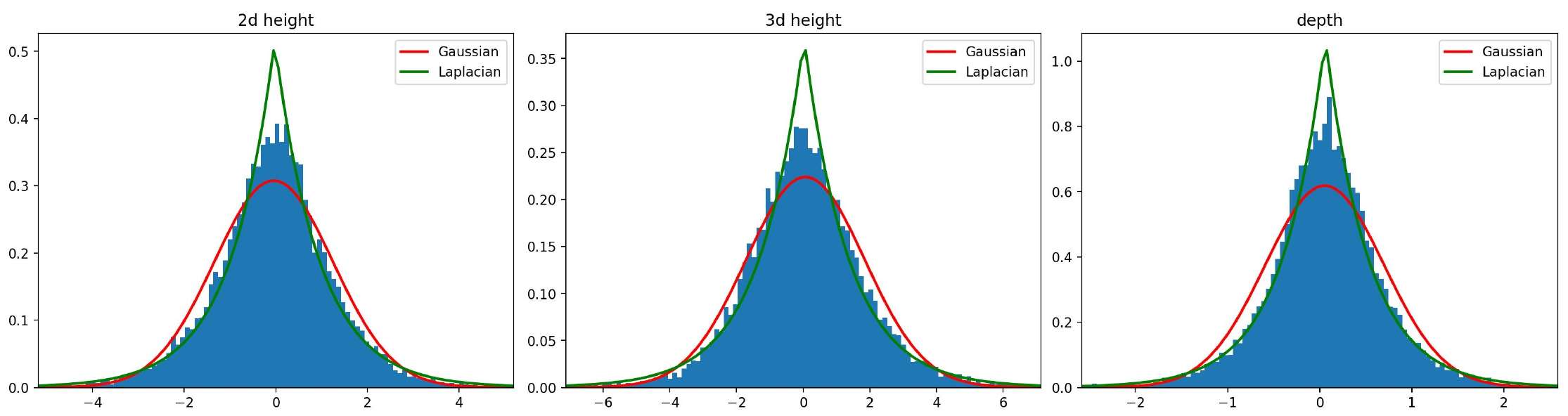}
    \caption{Histogram and fitting contours of depth, 2D height and 3D height distributions, respectively. }
    \label{fig:dist}
\end{figure*}

\noindent
2). We visualize the distribution forms for 2D height, 3D height, and depth to further support our choice of the Laplacian assumption. This visualization process is not straightforward. We provide an example of visualizing the depth distribution, as the visualization process for 2D and 3D heights follows the same methodology. This process is conducted on the training set. For each training sample, we apply the trained model to infer output depth information, which includes $\mu_d$ and $\sigma_d$. We then de-normalize the ground-truth depth $\hat{d}$ of the sample using the formula: $\frac{\hat{d}-\mu_d}{\sigma_d}$. Finally, we create a statistical histogram of $\frac{\hat{d}-\mu_d}{\sigma_d}$ across the entire training set to visualize the depth distribution. The results are shown in Figure~\ref{fig:dist}.

It could be seen that the Laplacian distribution (the \textcolor[RGB]{36,128,2}{green} line) could fit contours of statistical histograms more accurately than the Gaussian one (the \textcolor[RGB]{233,0,0}{red} line). Quantitatively, we calculate the average fitting errors for both two distributions in Table~\ref{tab:dist}. The Laplacian distribution shows significantly lower fitting errors than the Gaussian for Depth and 2D height, while achieving similar fitting gaps for 3D height. These results support our choice of Laplacian distribution in both qualitative and quantitative ways. 

\begin{table}[!t]
\begin{center}
\fontsize{8}{10}\selectfont
\caption{Mean fitting errors of Gaussian and Laplacian distributions for depth, 2D height and 3D height histograms, respectively.}
\label{tab:dist}
\begin{tabular}{l||ccc}
\toprule
\multirow{2}{*}{Distributions} & \multicolumn{3}{c}{Mean errors}\\ 
\cline{2-4} 
 ~ & Depth & 2D height & 3D height\\ 
\hline
Gaussian
& 0.01352 & 0.00787 & {\bf 0.00638} \\\
Laplacian
& {\bf 0.00698} & {\bf 0.00652} & 0.00614 \\  
\bottomrule
\end{tabular}
\end{center}
\end{table}

Finally, we provide the insight behind such a visualization process. Under the example of depth visualization, our goal is to visualize the form of the modeled distribution $p(d|img)$. There are two important tips should be noted:
\begin{itemize}
    \item We cannot directly count the depth in the training set as these statistics correspond to $p(d)$ rather than $p(d|img)$. 
    \item Our visualization target is the modeled $p(d|img)$ rather than the real $p(d|img)$ distribution because the latter is unknowable.
\end{itemize}
Visualizing $p(d|img)$ is challenging because the distribution of $d$ varies with different input images $img$. Fortunately, our goal is to illustrate the form of $p(d|img)$, whether Gaussian or Laplacian. Thus, we express $p(d|img)$ as follows:
\begin{equation}
    d = \mu_d+\sigma_d*\epsilon,\ \text{where}\ \{\mu_d,\sigma_d\}=f(img).
\end{equation}
where $f$ is the trained model and $\epsilon$ means a standard distribution. This kind of representation is well-used in the probabilistic model~\cite{bishop2006pattern}. Obviously, $\epsilon$ has the same form/type with the $p(d|img)$, so visualization of $\epsilon$ could show the form of $p(d|img)$. So, we create a statistical histogram of $\frac{\hat{d}-\mu_d}{\sigma_d}$ across the entire training set to show the depth distribution form.

\section{More visualization}
We provide additional visualization of the GUPNet++ about occlusion cases, comparison with state-of-the-art method and failure cases. 

\begin{figure*}[!ht]
    \centering
    \includegraphics[width=1.0\linewidth]{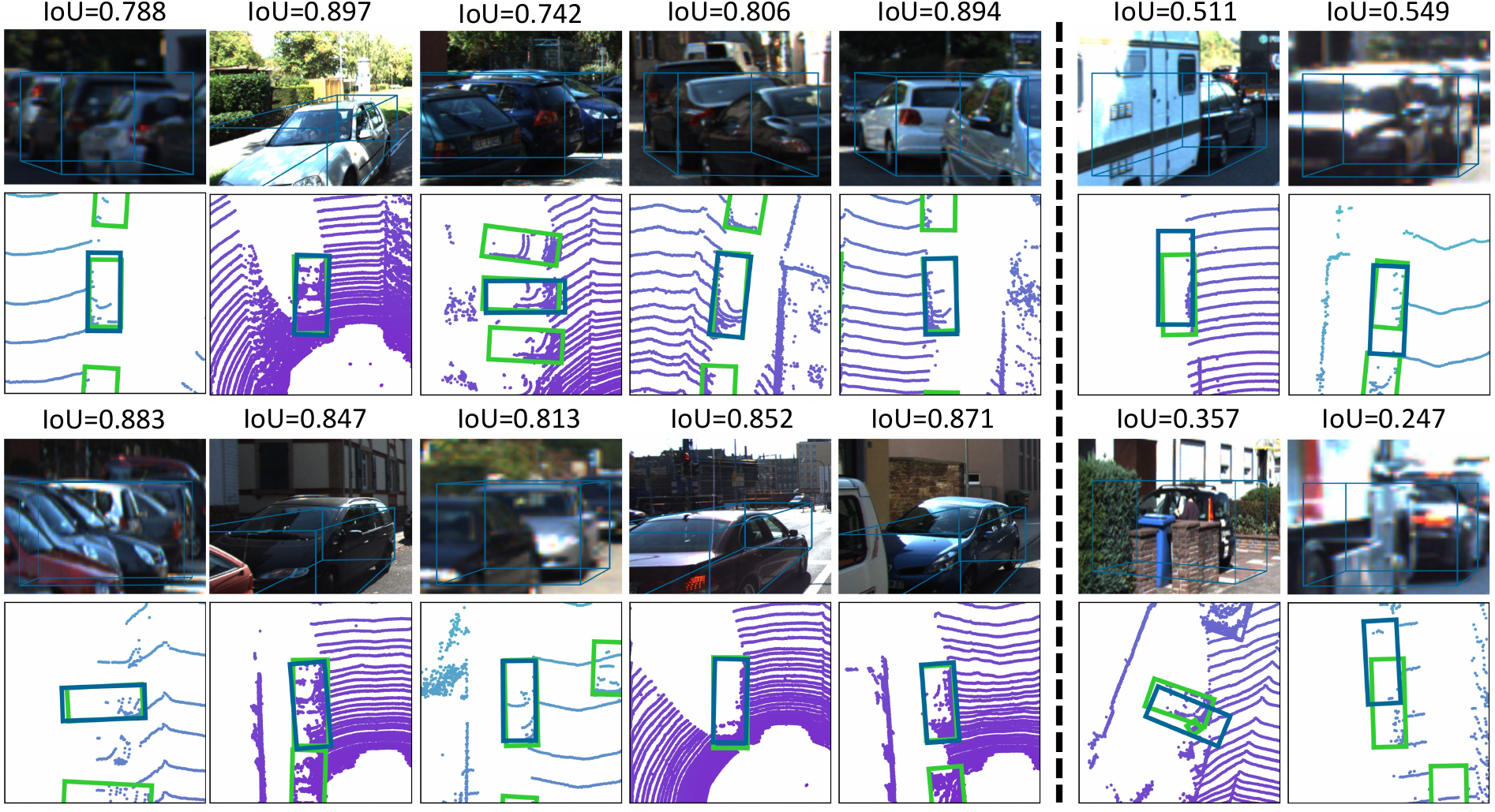}
    \caption{Visualization of occluded and truncated examples. Images to the left of the dashed line are good cases. And samples at the right are bad cases. \textcolor[RGB]{0,102,153}{Blue} boxes represents the box prediction and \textcolor[RGB]{50,205,50}{Green} boxes are the ground-truth boxes. }
    \label{fig:app_occ_cases}
\end{figure*}

{\bf Occlusion cases:}
We visualize some examples of occluded and truncated objects in Figure~\ref{fig:app_occ_cases}. It could be seen that the GUPNet++ works well for most occluded and truncated cases. However, for some extreme occlusion cases, the depth prediction may have certain noise. 

\begin{figure*}[!ht]
    \centering
    \includegraphics[width=1.0\linewidth]{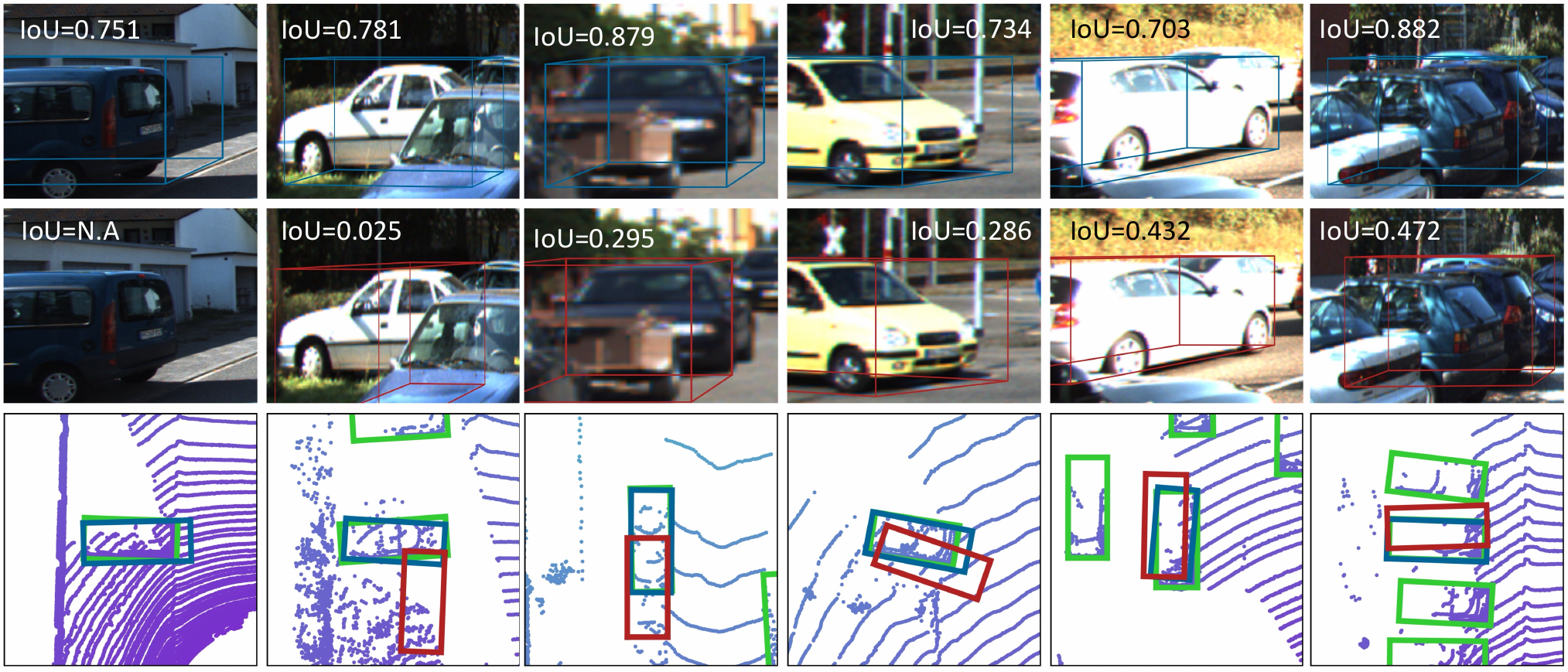}
    \caption{Qualitative comparison with MonoDETR~\cite{zhang2022monodetr}. The first row (\textcolor[RGB]{0,102,153}{Blue} boxes) are the results of our GUPNet++ while the second row (\textcolor[RGB]{178,34,34}{Red} boxes) are the MonoDETR results. The 3rd row shows the bird-view results (\textcolor[RGB]{50,205,50}{Green} means the ground truth boxes).}
    \label{tab:sup_monodetr}
\end{figure*}

{\bf Comparison with state-of-the-art method:} We provide additional qualitative comparison with state-of-the-art MonoDETR~\cite{zhang2022monodetr} as Figure~\ref{tab:sup_monodetr} shows. Compared with MonoDETR, our GUPNet++ could treat occlusion and truncation better, whose conclusion is consistent with the qualitative analysis in 2). It further demonstrates the occlusion- and truncation-robustness of our GUPNet++ 

{\bf Failure cases:} We visualize some failure cases in Figure~\ref{fig:app_fail_cases}. It can be seen that for some extremely low-quality imaged, small or occluded objects. The GUPNet++ would fail to detect them. 

\section{Limitations and future works}
We provide limitations of the GUPNet++ and potential future works as follows:

{\bf Limitations:}
The depth propagation process still has simplification: The 2D height $H_{2d}$ and 3D height $H_{3d}$ are both empirically assumed as Laplacian distribution. At this time, the distribution of the projected depth $D_{p}$ cannot written as a simple analytical formulation. Assuming $D_{p}$ as a Laplacian distribution is a simplification here, affecting the accuracy of depth uncertainty.

\begin{figure*}[!h]
    \centering
    \includegraphics[width=1.0\linewidth]{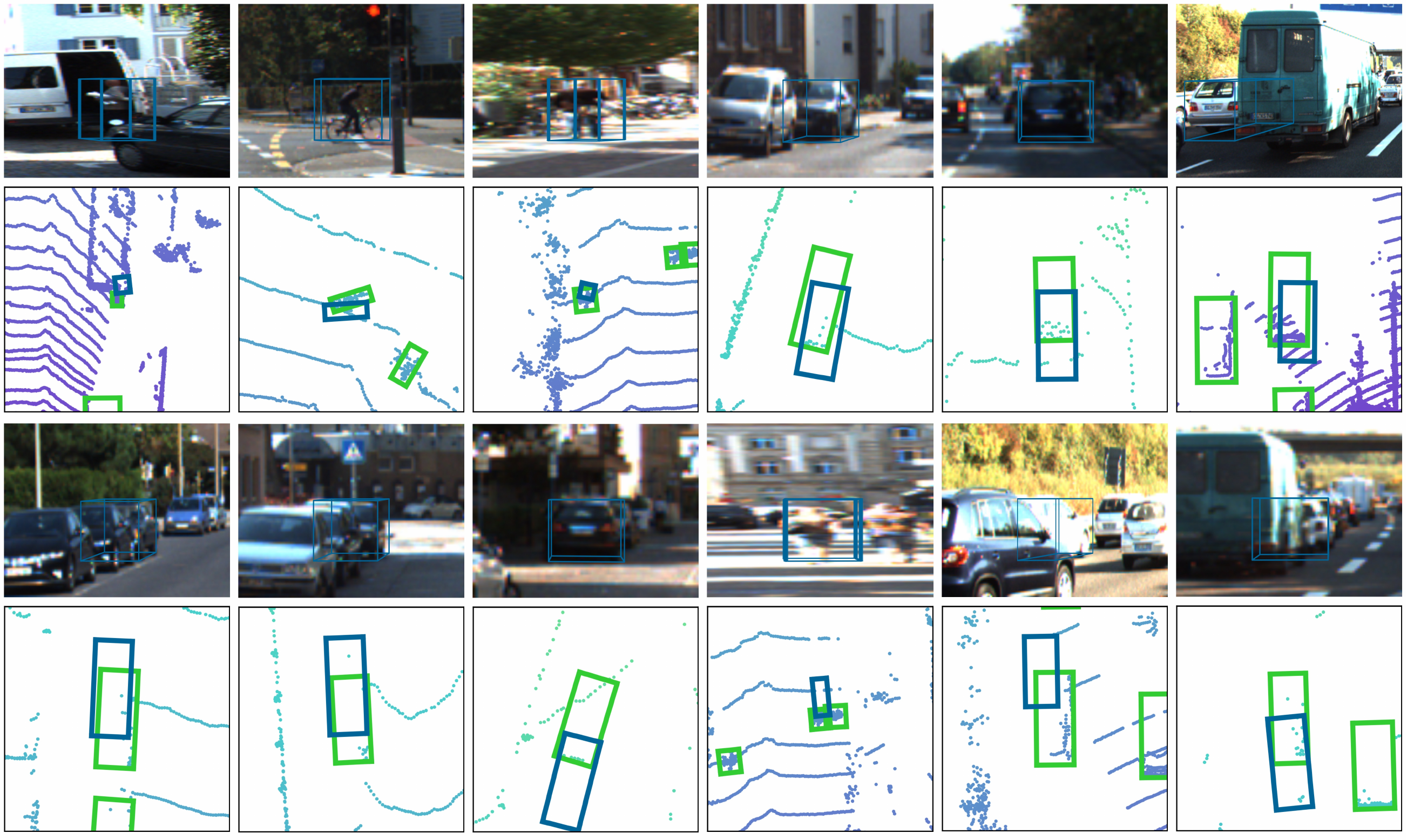}
    \caption{Visualization of some failure cases of the GUPNet++. \textcolor[RGB]{0,102,153}{Blue} boxes represents the box prediction and \textcolor[RGB]{50,205,50}{Green} boxes are the ground-truth boxes. }
    \label{fig:app_fail_cases}
\end{figure*}

{\bf Potential future works:}
\begin{enumerate}
\item For depth simplification, future works may combine advanced mathematical tools or generative models, such as flow model~\cite{li2021human}, to represent more complex depth distribution formulations, which may lead to better results.  
\item From our visualized failure cases in Figure~\ref{fig:app_fail_cases}, we could find that although the 3D bounding boxes are not accurate, their projected visual 2D boxes all seem good. Similar phenomena could be found in Figure 5 in the original paper. Different quality 3D boxes may lead to similar visual 2D projected boxes. This implies the error amplification in a different bounding box view, which may lead to other works. 
\end{enumerate}

\section{Geometry Uncertainty Projection (GUP)}
\label{sec:gup}
\subsection{Perspective Projection with Uncertainty Modeling}
\label{sec:geu_v1}
To model the perspective projection in a probabilistic manner, we assume that the predicted 3D height is a random variable rather than a fixed value. Then, the projection process is written as:
\begin{equation}
\label{eq:gup}
\begin{aligned}
    D_p=\frac{f\cdot H_{3d}}{h_{2d}},
\end{aligned}
\end{equation}
where $D_p$ and $H_{3d}$ mean projected depth and 3D height variables, respectively. To simplify the problem, $h_{2d}$ is assumed as a fixed value. 

Specifically, $H_{3d}$ is assumed following Laplacian distribution $La(\mu_{3d},\sigma_{3d})$ where its mean $\mu_{3d}$ and standard deviation (std) $\sigma_{3d}$ are both predicted by the 3D size head. Based on the modeled $h_{3d}$ distribution, $D_p$ is derived as:
\begin{equation}
\label{eq:gupv1}
\begin{aligned}
    D_{p} &= \frac{f\cdot H_{3d}}{h_{2d}}
    = \frac{f\cdot (\sigma_{3d}\cdot X + \mu_{3d})}{h_{2d}}\\
    & = \frac{f\cdot \sigma_{3d}}{h_{2d}}\cdot X + \frac{f\cdot \mu_{3d}}{h_{2d}},
\end{aligned}
\end{equation}
where $X$ is the standard Laplacian random variable. In this sense, the mean and std of the projected depth are:
\begin{equation}
\label{eq:geu_appendix}
\begin{aligned}
    \mu_{p} &= \frac{f\cdot \mu_{3d}}{h_{2d}},\ \ \ 
    \sigma_{p} = \frac{f\cdot \sigma_{3d}}{h_{2d}}. 
\end{aligned}
\end{equation}
As referred in APPENDIX~\ref{sec:app_depth_bias} and to acquire better depth. GUPNet also utilizes a biased depth modification scheme as follows:
\begin{equation}
\label{eq:bias_appendix}
\begin{aligned}
     \mu_d = \mu_{p}+\mu_{b},\quad \sigma_{d} &= \sqrt{(\sigma_{p})^2+(\sigma_{b})^2}.
\end{aligned}
\end{equation}
Here, we refer to the $\sigma_{d}$ as Geometry-guided Uncertainty (GeU).

\subsection{Inference with vanilla Uncertainty-Confidence}
\label{sec:EUS}
In the GUP module, $p_{3d|2d}$ is computed as following:
\begin{equation}
\label{eq:expus}
\begin{aligned}
     p_{3d|2d} = exp(-\sigma_{d}).
\end{aligned}
\end{equation}
We call this scheme the vanilla Uncertainty-Confidence (UnC). It utilizes an exponential function to map the depth uncertainty $\sigma_{d}$ to a value between 0$\sim$1 making the output can be seen as a probability. With such UnC, an unreliable depth prediction with high uncertainty will have low confidence and vice versa. 

\subsection{Training with Hierarchical Task Learning}
\label{sec:htl}
During training, we use L1 loss for 2D height:
\begin{equation}
\begin{aligned}
    \mathcal{L}_{h2d} = |h_{2d}-\hat{h}_{2d}|.
\end{aligned}
\end{equation}
For training the depth and 3D height distributions, we utilize the standard Laplacian uncertainty regression loss, also called Negative Log-Likelihood (NLL) loss:
\begin{equation}
\label{eq:depth_loss}
\begin{aligned}
\mathcal{L}_{h3d} =
    \frac{\sqrt{2}}{\sigma_{3d}}|\mu_{3d}-\hat{h}_{3d}|+log(\sigma_{3d})\\
    \mathcal{L}_{depth} = \frac{\sqrt{2}}{\sigma_{d}}|\mu_{d}-\hat{d}|+log(\sigma_{d}).
\end{aligned}
\end{equation}
This loss function trains the $\mu$ and $\sigma$ in an end-to-end manner. With this uncertainty loss function, a sample with a high $\sigma$ value will have a decreased sample weight
, leading to better training efficiency for ill-posed tasks, 3D size and depth estimation. 


Further, to address the training instability caused by the error amplification effect in the training stage. GUP module proposes a curriculum learning scheme, Hierarchical Task Learning (HTL), to train the different loss terms in a specific order. For a loss term $L_i$, HTL controls its loss weight at each epoch. The overall loss is organized as follows:
\begin{equation}
\begin{aligned}
     \mathcal{L}_{total} = \sum_{i\in \mathcal{T}} w_{i}(t)\cdot \mathcal{L}_{i},
\end{aligned}
\label{eq:app_htl_loss}
\end{equation}
where $\mathcal{T}$ is the task set including \{heatmap, 2D size, 2D offset, 3D size, 3D offset, angle, depth\}. The derivation of the $w_{i}(t)$ follows the idea that each task should commence training after its predecessor tasks have been well-trained. So, in HTL, we split tasks into different stages and the loss weight $w_{i}(t)$ is associated with all pre-tasks of the $i$-th task. With that, depth is trained after its pre-tasks, two height estimations, achieving satisfying training situations and making the whole training process of the perspective projection stable. Details can be seen in the conference paper~\cite{lu2021geometry} or APPENDIX~\ref{sec:app_htl}.

\section{Hierarchical Task Learning}
\label{sec:app_htl}
The GUP module mainly addresses the error amplification effect in the inference stage. Yet, this effect also damages the training procedure. Specifically, at the beginning of the training, the prediction of both $h_{2d}$ and $h_{3d}$ are far from accurate, which will mislead the overall training and damage the performance.  
To tackle this problem, we design a Hierarchical Task Learning (HTL) to control weights for each task at each epoch. The overall loss is shown in Equation~\ref{eq:app_htl_loss}.

\begin{figure}[t]
\begin{center}
\includegraphics[width=1.0\linewidth]{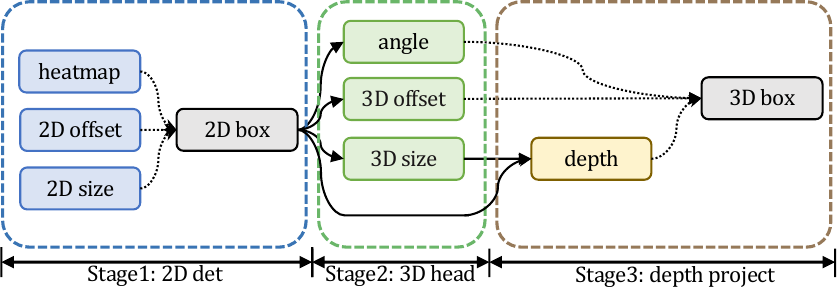}
\end{center}
   \caption{The task hierarchy of the GUP Net. The first stage is
2D detection. Built on top of RoI features, the second stage
consists of basic 3D detection heads. Based on 2D and 3D heights
estimated in the previous stages, the third stage infers the depth and
then constitutes the 3D bounding box.}
\label{fig:task_graph}
\vspace{-0.3cm}
\end{figure}

\begin{figure}[t]
\begin{center}
\includegraphics[width=0.8\linewidth]{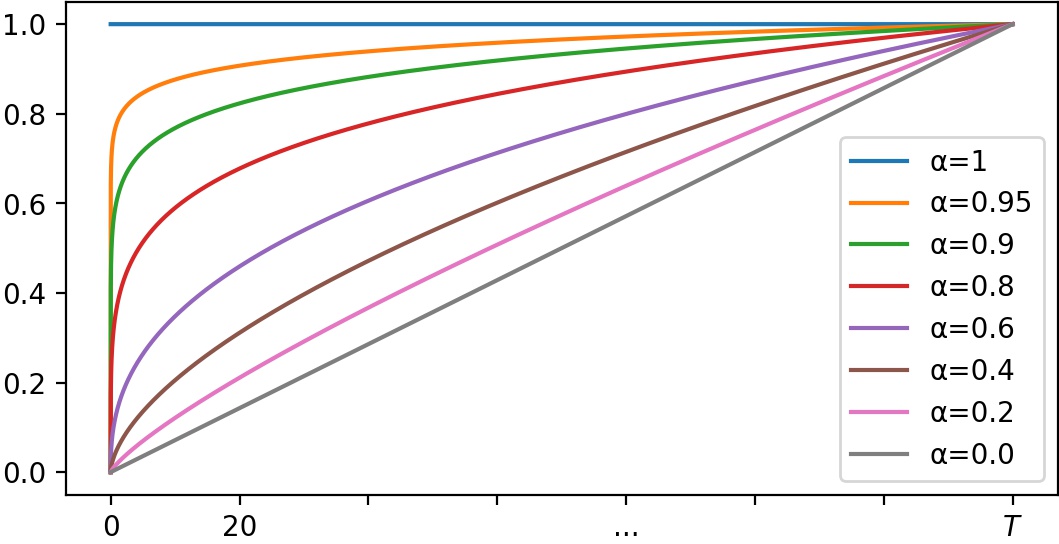}
\end{center}
   \caption{The polynomial time scheduling function with the adjust parameter. The vertical axis is the value of $w_i(t)$ and the horizontal axis is the epoch index $t$. (best viewed in color.)}
\label{fig:time_schedule}
\end{figure}

HTL is inspired by the motivation that each task should start training after its pre-task has been trained well. We split tasks into different stages as shown in Figure~\ref{fig:task_graph} and the loss weight $w_{i}(t)$ should be associated with all pre-tasks of the $i$-th task. The first stage is 2D detection, including heatmap, 2D offset, 2D size. Then, the second stage is the 3D heads containing angle, 3D offset and 3D size. All of these 3D tasks are built on the ROI features, so the tasks in 2D detection stage are their pre-tasks. Similarly, the final stage is the depth inference and its pre-tasks are the 3D size and all the tasks in 2D detection stage since depth prediction depends on the 3D height and 2D height.
To train each task sufficiently, we aim to gradually increase the $w_{i}(t)$ from 0 to 1 as the training progresses. So we adopt the widely used polynomial time scheduling function~\cite{morerio2017curriculum} in the curriculum learning topic as our weighted function, which is adapted as follows:
\begin{equation}
\begin{aligned}
     w_{i}(t) = (\frac{t}{T})^{1-{\alpha_i(t)}},\ \ \alpha_i(t) \in [0,1],
\end{aligned}
\end{equation}
where $T$ is the total training epochs and the normalized time variable $\frac{t}{T}$ can automatically adjust the time scale. ${\alpha_i}(t)$ is an adjust parameter at the $t$-th epoch, corresponding to every pre-task of the $i$-th task. Figure~\ref{fig:time_schedule} shows that ${\alpha_i}$ can change the trend of the time scheduler. The larger ${\alpha_i}$ is, the faster $w_{i}(\cdot)$ increases.
From the definition of the adjust parameter, it is natural to decide its value via the learning situation of every pre-task. If all pre-task have been well trained, the ${\alpha_i}$ is expected be large, otherwise it should be small. This is motivated by the observation that human usually learn advanced courses after finishing fundamental courses. Therefore, ${\alpha_i}(t)$ is defined as:
\begin{equation}
\begin{aligned}
     {\alpha_i}(t) = \prod_{j\in{\mathcal{P}_i}}ls_j(t),
\end{aligned}
\end{equation}
where $\mathcal{P}_i$ is the pre-task set for the $i$-th task. $ls_j$ means the learning situation indicator of $j$-th task, which is a value between 0$\sim$1. This formula means that $\alpha_i$ would get high values only when all pre-tasks have achieved high $ls$ (trained well). For the $ls_j$, inspired by~\cite{chen2018gradnorm,zhang2014facial}, we design a scale-invariant factor to indicate the learning situation:
\begin{equation}
\begin{aligned}
     {ls_j}(t) &= \frac{\mathcal{DF}_j(K)-\mathcal{DF}_j(t)}{\mathcal{DF}_j(K)},\\
     \mathcal{DF}_j(t) &= \frac{1}{K}\mathop{\sum}\limits_{\hat{t}=t-K}^{t-1}|\mathcal{L}'_j\left( \hat{t}\right)|,
\end{aligned}
\end{equation}
where $\mathcal{L}'_j(\hat{t})$ is the derivative of the $\mathcal{L}_j(\cdot)$ at the $\hat{t}$-th epoch, which can indicate the local change trend of the loss function. The $\mathcal{DF}_j(t)$ computes the mean of derivatives in the recent $K$ epochs before the $t$-th epoch to reflect the mean change trend. If the $\mathcal{L}_j$ drops quickly in the recent $K$ epochs, the $\mathcal{DF}_j$ will get a large value. So the ${ls_j}$ formula means comparing the difference between the current trend $\mathcal{DF}_j(t)$ and the trend of the first $K$ epochs at the beginning of training $\mathcal{DF}_j(K)$ for the $j$-th task. If the current loss trend is similar to the beginning trend, the indicator will give a small value, which means that this task has not been trained well. Conversely, if a task tends to converge, the ${ls_j}$ will be close to 1, meaning that the learning situation of this task is satisfied.
Based on the overall design, the loss weight of each term can reflect the learning situation of its pre-tasks dynamically, which can make the training more stable.

\end{document}